\documentclass{article}

\PassOptionsToPackage{numbers,sort&compress}{natbib}
\usepackage[preprint]{neurips_2026}
\usepackage[utf8]{inputenc}
\usepackage[T1]{fontenc}
\usepackage{hyperref}
\usepackage{url}
\usepackage{booktabs}
\usepackage{amsfonts}
\usepackage{amsmath}
\usepackage{amssymb}
\usepackage{microtype}
\usepackage[final]{graphicx}
\usepackage{float}
\usepackage{placeins}
\usepackage{wrapfig}
\usepackage{multirow}
\usepackage{tabularx}
\usepackage{xcolor}
\graphicspath{{diagnostic_analysis_materials/}}
\setkeys{Gin}{draft=false}
\title{Will Accurate Fields Mislead Photonic Design? From Global Accuracy to Port Readout}

\author{
Yitian Zhang$^{1}$, Yonghong Chen$^{1}$, Youming Chen$^{1}$,
Yiyang Li$^{1}$, Zhe Xing$^{1}$\\
Rehen Lu$^{1}$, Shaolin Liao$^{1}$,
Yuzhe Ma$^{2}$, Zhong Guan$^{1,*}$\\[0.5em]
{\small $^{1}$ Sun Yat-sen University}\\
{\small $^{2}$ The Hong Kong University of Science and Technology (Guangzhou)}\\
{\small $^{*}$ Corresponding Author}
}

\providecommand{\answerYes}[1][]{\textcolor{blue}{[Yes] #1}}
\providecommand{\answerNo}[1][]{\textcolor{orange}{[No] #1}}
\providecommand{\answerNA}[1][]{\textcolor{gray}{[NA] #1}}
\begin{document}

\maketitle
\pagenumbering{arabic}
\setcounter{page}{1}
\renewcommand{\thepage}{\arabic{page}}

\begin{abstract}
Neural field surrogates can accelerate photonic design loops, but a surrogate that looks accurate in global field error can still mis-rank candidate devices when the final decision depends on localized output-port readouts.
This risk is acute in propagation-dominated MMI splitters and couplers, where port power, splitting, phase, and coupling are determined by accumulated modal interference and output-window aggregation rather than by average field similarity alone.
We study this field-to-design mismatch through a Field/Mediator/Readout view that separates dense complex-field error from propagation-profile and output-window errors before port aggregation.
To align the surrogate with this chain, we propose PaNO, a propagation-aligned neural operator that keeps the full-field prediction interface while organizing latent states around local boundary structure, transverse modal content, axial propagation, and cross-mode interaction.
We also evaluate PaNO-R2, an output-aware feedback variant for residual field components near the port region.
On a 15-wavelength tunable $3{\times}3$ MMI benchmark with 4608 held-out fields, PaNO lowers NeurOLight's port-power error from 0.2018 to 0.0739 despite slightly higher cMAE, showing that global field accuracy alone is not sufficient for design-relevant readout fidelity.
PaNO-R2 attains the best cMAE, propagation-profile error, output-profile error, and port-power error, reducing NeurOLight's port-power and output-profile errors by 72.7\% and 72.5\%.
\end{abstract}

\section{Introduction}

Photonic devices are central to optical communication, on-chip optical networks, and optical computing \citep{miller2013selfconfiguring,bogaerts2020programmable,shastri2021photonics_ai}.
Designing these devices requires repeated electromagnetic simulation over geometries, wavelengths, and input excitations, which makes high-fidelity FDFD/FDTD-style solvers expensive inside parameter sweeps or inverse-design loops \citep{taflove2005fdtd,oskooi2010meep,molesky2018inverse_design}.
Neural field surrogates offer a practical alternative: they predict the complex optical field once trained, reducing the cost of evaluating many candidate devices \citep{li2021fno,lu2021deeponet,gu2022neurolight}.

The difficulty is that photonic design decisions are usually made from \textbf{device readouts}, not from global field similarity alone.
For MMI splitters and couplers, designers inspect output-port powers, splitting ratios, relative phases, and coupling behavior \citep{soldano1995mmi_self_imaging,bachmann1994mmi_phase,hill2003mmi_imbalance}.
Dense complex-field errors such as cMAE remain important because every readout is computed from the predicted field.
However, they average over the computational window, whereas port quantities are localized at output windows and depend on coherent modal interference.
A surrogate can therefore look accurate in the full field while still producing a wrong port readout, which can \textbf{mis-rank candidate layouts} in a sweep or inverse-design loop.
The central question of this paper is whether field accuracy alone diagnoses the \textbf{propagation-to-readout chain} that matters for device selection.

\begin{figure}[H]
  \centering
  \includegraphics[width=\columnwidth]{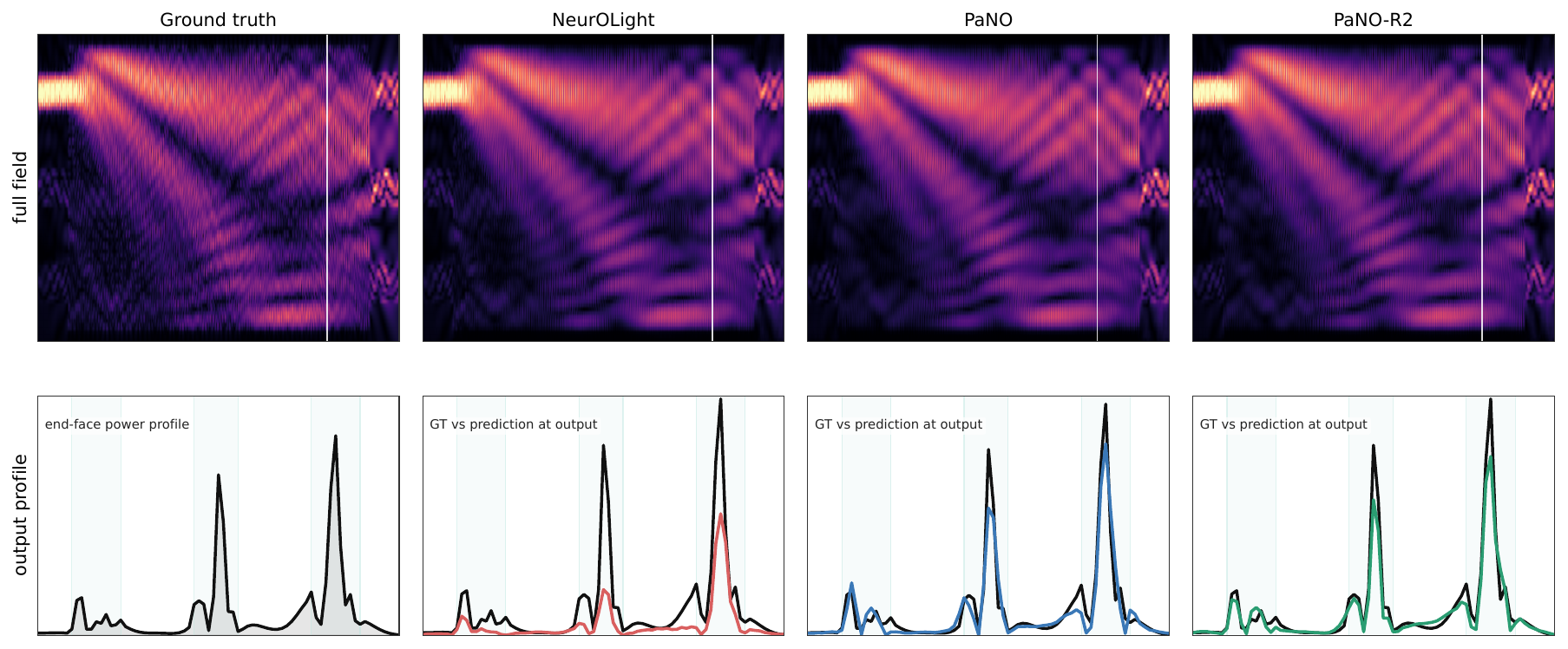}
  \caption{Opening example of the field/readout mismatch. Columns compare ground truth, NeurOLight, PaNO, and PaNO-R2; rows show full-field amplitude and the output-plane power profile used for port integration. Broad field similarity can still hide localized output-profile shifts near the port windows, while PaNO and PaNO-R2 better align the readout region.}
  \label{fig:problem-diagnosis}
\end{figure}

MMI devices make this mismatch especially concrete because their output profiles are produced by modal phases and interference accumulated along the propagation axis, not by a local texture near the port \citep{snyder1983optical_waveguide,soldano1995mmi_self_imaging,bachmann1994mmi_phase}.
Figure~\ref{fig:problem-diagnosis} illustrates the failure mode: all models retain the broad propagation pattern, yet the output-plane intensity profile inside the port windows can shift substantially.
The example is motivational; the experiments below test whether this field/readout gap is systematic.

We address this question with a Field/Mediator/Readout view of photonic surrogates.
Field metrics measure dense complex-field reconstruction, Mediator metrics measure propagation and output-profile consistency before port aggregation, and Readout metrics measure localized device quantities.
This decomposition keeps cMAE as a necessary field-fidelity metric, but adds the intermediate quantities that connect full-field prediction to the device functions used in design.
It also avoids reducing the problem to a scalar port predictor: \textbf{all models remain full-field surrogates}, so the predicted complex field can still support inspection, multiple readouts, and physics diagnostics.
In this work we instantiate this view on tunable MMI devices with localized output ports.

We also propose PaNO, a \textbf{propagation-aligned neural operator} for this setting.
PaNO remains a full-field surrogate with no separate scalar port head, but its latent computation follows the device physics more closely than a generic image-like operator.
It encodes local boundary and output-window structure, forms learned transverse modal tokens, propagates them along the physical axis with a selective state-space scan, and couples modes before decoding.
We further evaluate PaNO-R2, an output-sensitive reverse-residual variant, to test whether global field fidelity and output-side mediator accuracy can be improved together.

On the 15-wavelength MMI evaluation, PaNO improves SWR, propagation profile, output profile, and port power over NeurOLight despite slightly worse cMAE.
PaNO-R2 gives the best cMAE, SWR, propagation profile, output profile, port power, and splitting among completed rows, while phase and coupling remain mixed.
The evidence supports a bounded conclusion: for propagation-dominated photonic devices with localized readouts, surrogates should be evaluated and designed around the \textbf{full Field/Mediator/Readout chain} rather than dense-field error alone.

\textbf{The specific contributions of this paper are as follows:}
\emph{First,} we empirically identify a proxy-objective mismatch between dense-field cMAE and device-level photonic readouts.
\emph{Second,} we formulate a Field/Mediator/Readout diagnostic view, supported by a simple port-power bound that connects output-window intensity profiles to localized readout error.
\emph{Third,} we propose PaNO, a propagation-aligned neural operator with local encoding, modal sequence propagation, and cross-mode coupling.
\emph{Fourth,} we validate the diagnosis and architecture through 15-wavelength MMI comparisons and ablations that show both improvements and trade-offs.

\section{Related Work}

Neural operators model mappings from input conditions to output fields and are now standard surrogates for PDE-governed systems.
FNO, DeepONet, and later variants establish this full-field interface for function-space prediction \citep{li2021fno,lu2021deeponet,kovachki2023neural_operator,cao2021galerkin_transformer,hao2023gnot}, and NeurOLight brings the same paradigm to Maxwell photonic simulation \citep{gu2022neurolight}.
We keep this interface, but focus on a mismatch that is easy to miss in dense reconstruction: photonic design often selects devices by localized port powers, splitting ratios, phases, and coupling quantities rather than by global field error alone \citep{lalaukeraly2013adjoint,molesky2018inverse_design,liu2018inverse_dnn}.
For MMI splitters and couplers, these readouts are produced by coherent multimode interference and localized output windows \citep{soldano1995mmi_self_imaging,bachmann1994mmi_phase,hill2003mmi_imbalance}.
Thus a visually plausible field can still shift the output profile that a port integrates.
Our Field/Mediator/Readout evaluation keeps all models as full-field predictors, but tests whether their fields preserve the propagation-to-readout chain.

This readout sensitivity also motivates a propagation-structured architecture.
Classical photonic modeling exploits axial propagation and modal organization through BPM and EME-style views \citep{snyder1983optical_waveguide,feit1978graded_index_bpm,bienstman2001eme_pml}, while state-space sequence models offer learnable long-range dynamics \citep{gu2022s4,nguyen2022s4nd,gu2024mamba}.
PaNO uses these ideas as inductive bias rather than as a solver: it learns transverse modal tokens, selective axial dynamics, and cross-mode interactions, then decodes a full complex field.
The contribution is therefore not a new electromagnetic discretization or a separate port head, but a propagation-aligned surrogate and diagnostic protocol for localized photonic readouts.

\section{Problem Setting and Motivation}
\label{sec:problem}

\subsection{Frequency-Domain Port Readout}

We study full-field surrogate modeling for two-dimensional, Hz-polarized multimode-interference devices with localized output ports \citep{soldano1995mmi_self_imaging,bachmann1994mmi_phase}.
For a relative-permittivity map $\varepsilon_r(y,w)$, wavelength $\lambda$, and input-port excitation $s_{\mathrm{in}}$, the steady-state magnetic field $u=H_z\in\mathbb{C}^{H\times W}$ is the solution of a scalar frequency-domain Helmholtz/FDFD system,
\begin{equation}
  A_{\omega,\varepsilon}u=b(s_{\mathrm{in}}),
  \qquad
  \omega=\frac{2\pi c}{\lambda},
  \label{eq:helmholtz-fdfd}
\end{equation}
where $A_{\omega,\varepsilon}$ is the discretized Maxwell operator determined by the device geometry, material distribution, boundary condition, and wavelength.
The learning task is to amortize this solve: the model receives a grid-aligned embedding of the geometry and source conditions and predicts the full complex field,
\begin{equation}
  x=\mathrm{Embed}(\varepsilon_r,s_{\mathrm{in}},\lambda),
  \qquad
  \hat E \equiv \hat u=f_\theta(x)\in\mathbb{C}^{H\times W}.
  \label{eq:field-prediction}
\end{equation}
All device metrics are then computed from $\hat u$ by fixed readout operators; no separate scalar port head is used.

Port power is a localized output-window readout of this predicted field.
In full electromagnetic simulation tools, port power is commonly evaluated by Poynting flux or by modal-overlap decompositions at waveguide cross sections, as in standard Meep- or Ansys-style port analyses \citep{oskooi2010meep,fallahkhair2008modesolver}.
Our evaluator uses the corresponding scalar Hz-field intensity proxy for the fixed port mask $m_p(x)\in[0,1]$:
\begin{equation}
  P_p(u)=\sum_{x\in\Omega_p}m_p(x)|u(x)|^2,
  \label{eq:port-power}
\end{equation}
where $\Omega_p$ denotes the support of the localized output-window mask for port $p$.
This definition makes port power a local functional of the output intensity envelope rather than a spatially uniform average over the whole simulated field.
The intensity proxy is the readout analyzed in the next subsection; phase and coupling metrics additionally depend on coherent complex-field structure and are therefore not fully characterized by port intensity alone.

\subsection{From Global Error to Propagation Readout}

Dense cMAE is a useful field-fidelity metric, but it answers a different question from localized port readout.
It averages error over the full computational window, whereas port power in Eq.~\eqref{eq:port-power} integrates intensity inside a small output mask.
Thus a model can improve average field error while still shifting the output-plane intensity envelope that the port integrates.
This is not a failure of cMAE as a field metric; it is a mismatch between a global reconstruction objective and a localized device readout.

For MMI devices, this mismatch is tied to propagation physics \citep{soldano1995mmi_self_imaging,bachmann1994mmi_phase,hill2003mmi_imbalance}.
In the standard self-imaging view, the field in the multimode region can be written as a superposition of transverse modes,
\begin{equation}
  u(y,w)\approx \sum_{m=1}^{M} c_m\phi_m(y)e^{i\beta_m w},
  \qquad
  P_p(u)\approx
  \int_{\Omega_p}
  m_p(y)
  \left|
  \sum_m c_m\phi_m(y)e^{i\beta_m L}
  \right|^2dy .
  \label{eq:mmi-propagation-readout}
\end{equation}
The output-window envelope is therefore not just a local texture near the port; it is the result of accumulated modal phases and interference along the propagation axis.
Small errors in transverse organization, self-imaging phase, or tail-end envelope can be modest in global cMAE but visible after port integration.

This motivates the experimental protocol used below.
We report \emph{Field} metrics for dense reconstruction, \emph{Mediator} metrics for propagation and output-envelope behavior before port aggregation, and \emph{Readout} metrics for localized device quantities.
The same locality also affects optimization: under a dense-field loss, small output windows receive limited direct pressure unless the model representation keeps propagation-to-output information active.
The next section instantiates this principle as a full-field neural operator with propagation-structured latent computation.

\section{Propagation-Aligned Neural Operator}
\label{sec:method}

The preceding analysis suggests a design goal beyond lowering dense cMAE: a useful photonic surrogate should preserve the propagation mediators from which localized port readouts are computed.
PaNO keeps the standard full-field interface, predicting $\hat E=f_\theta(x)$ from Eq.~\eqref{eq:field-prediction}, but aligns its latent computation with MMI propagation.
It uses MSAS for local boundary and output-window structure, learned transverse modal tokens for each propagation slice, a directed sequence backbone for axial transport, and residual cross-mode interaction with optional R2 compensation before decoding.

This organization is inspired by BPM- and EME-style photonic modeling, but it is used as a learnable inductive bias rather than a solver discretization \citep{feit1978graded_index_bpm,bienstman2001eme_pml,fallahkhair2008modesolver}.
PaNO does not require paraxial envelope updates, physical eigenmode solves, scattering-matrix assembly, or a separate port head; any readout improvement must come through the predicted complex field itself.

\begin{figure}[H]
    \centering
    \includegraphics[width=0.85\linewidth]{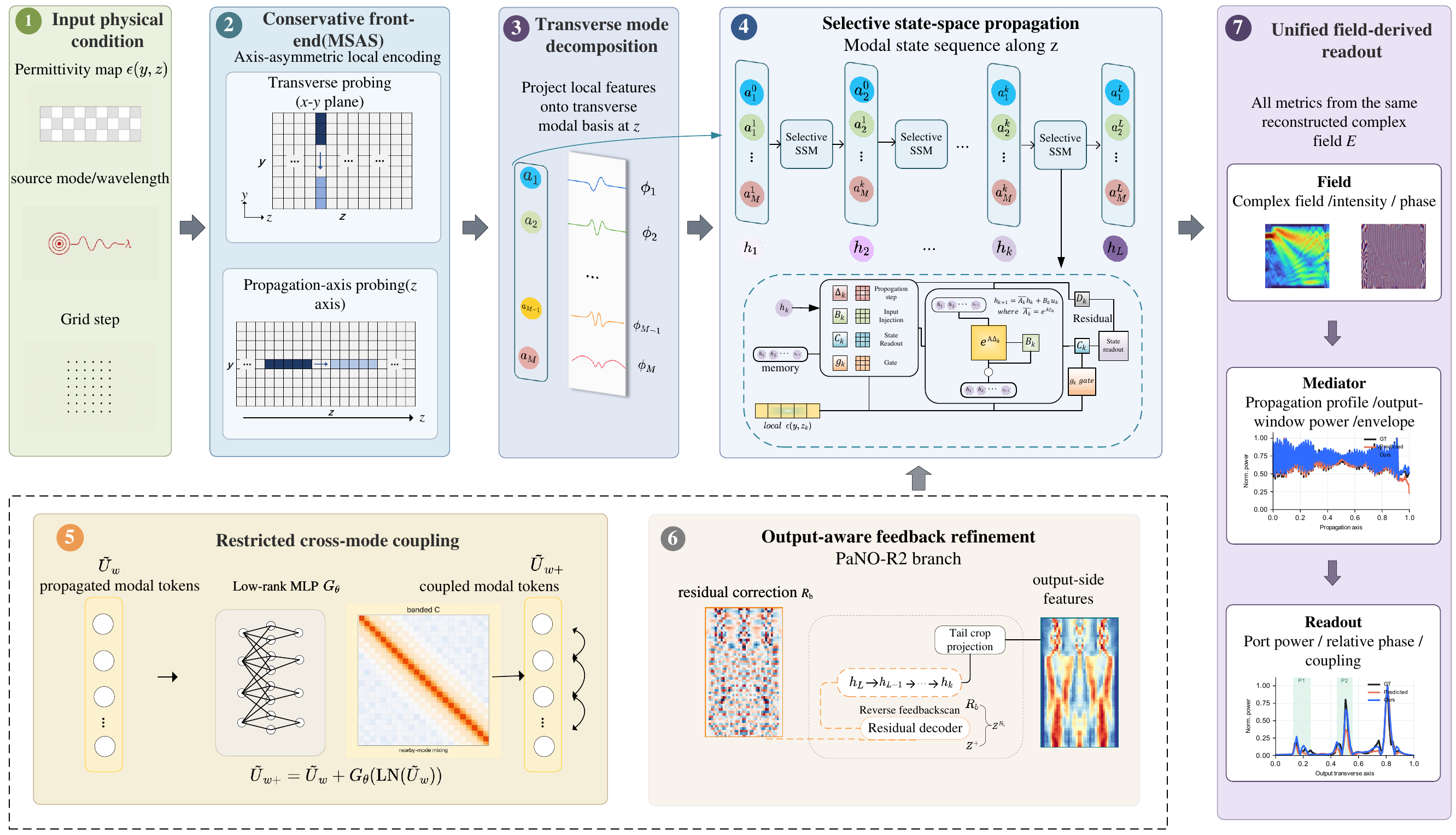}
    \caption{Propagation-aligned modal representation }
    \label{fig:pano-method-overview}
\end{figure}

\subsection{Multi-Scale Anisotropic Stem}

MSAS encodes the anisotropy of MMI fields before global mixing.
Along the propagation axis $w$, elongated interference envelopes call for wider local aggregation; along the transverse axis $y$, material boundaries and port-window edges call for shorter filters.
We therefore separate channel lifting, propagation-axis filtering, and transverse filtering:
\begin{align}
X_0 &= \mathrm{Conv}_{1\times 1}(X), \\
X_1 &= \mathrm{DWConv}_{1\times k_w}(X_0), \\
X_2 &= \mathrm{DWConv}_{k_y\times 1}(X_1), \\
X_3 &= \mathrm{Conv}_{1\times1}\!\left(\sigma\!\left(\mathrm{Conv}_{1\times1}(X_2)\right)\right), \\
X^0 &= \mathrm{GN}(X_3+X_0).
\label{eq:msas}
\end{align}
The $1\times k_w$ depthwise filter follows propagation envelopes along $w$, while the $k_y\times1$ filter detects transverse index jumps, waveguide edges, and output-window structure.
The residual normalization then produces conservative local features for the modal backbone.
This stem is a propagation-stability bias rather than a universal encoder: the ablations show that a less restricted pointwise/window stem can fit some in-distribution field metrics more tightly, while MSAS better preserves tail and output-profile behavior.

\subsection{Modal Sequence Backbone}

\subsubsection{Learned Neural Modal Decomposition}

Equation~\eqref{eq:mmi-propagation-readout} shows that port readout depends on transverse modal content at the output plane.
Rather than scanning raw image columns, PaNO maps each transverse slice into learned modal tokens,
\begin{equation}
  \mathbf{u}_{m,w}
  =
  W_{\mathrm{in}}
  \left(
  \sum_{y=1}^{H}\Phi_{m,y}X^0_{:,y,w}
  \right),
  \qquad
  m=1,\ldots,M ,
\label{eq:modal-token}
\end{equation}
where $X^0\in\mathbb{R}^{C\times H\times W}$ is the stem feature map.
The basis $\Phi$ is not a physical eigenmode basis; it is a learned transverse coordinate system that exposes modal organization while retaining end-to-end full-field supervision.

\subsubsection{Propagation as State-Space Stepping}

The modal representation also motivates a directed state update along the propagation axis.
Starting from Eq.~\eqref{eq:helmholtz-fdfd}, a local expansion $u(y,w)=\sum_m a_m(w)\phi_m(y;w)$ gives a coupled modal evolution (Appendix~\ref{app:modal-derivation}),
\begin{equation}
  \frac{d\mathbf a}{dw}
  =
  K_{\omega,\varepsilon(w)}\mathbf a(w)+\mathbf r(w),
  \qquad
  \mathbf a_{w+1}
  \approx
  T_{\omega,\varepsilon(w)}\mathbf a_w+\mathbf q_w .
\label{eq:modal-stepping}
\end{equation}
Here $\mathbf a_w$ is the physical modal-coefficient state, $T_{\omega,\varepsilon(w)}$ is the section-dependent propagation operator, and $\mathbf q_w$ collects source, boundary, truncation, and unmodeled-coupling effects.
PaNO keeps this state-update organization but replaces physical coefficients and transfer matrices with learned latent quantities.
It implements the propagation step as
\begin{align}
  \boldsymbol{\psi}_{m,w}
  &=
  \mathcal{S}_{\theta}(\mathbf{u}_{m,w},\mathbf{h}_{m,w}),\\
  \mathbf{h}_{m,w+1}
  &=
  \mathcal{T}_{\theta}(\mathbf{h}_{m,w},\mathbf{u}_{m,w};\boldsymbol{\psi}_{m,w}),\\
  \tilde{\mathbf{u}}_{m,w}
  &=
  \mathbf{u}_{m,w}
  +
  \mathcal{O}_{\theta}(\mathbf{h}_{m,w+1},\mathbf{u}_{m,w}).
\label{eq:ssm}
\end{align}
Here $\mathcal{S}_{\theta}$ selects local propagation parameters, $\mathcal{T}_{\theta}$ is the learned analogue of the homogeneous transfer, and $\mathcal{O}_{\theta}$ maps the updated state back to token space with a residual correction.
Thus $\tilde{\mathbf{u}}_{m,w}$ is the propagated modal token.
This data-dependent SSM adapts to local geometry and propagation context, giving PaNO a linear-complexity path to preserve self-imaging envelopes and tail-end profiles without acting as a traditional PDE solver.

\subsection{Controlled Modal Coupling and Reverse Residual Compensation}
\label{sec:r2}

The sequence scan primarily transports modal tokens independently, but localized port quantities are determined by coherent modal superposition at the output plane.
Writing a local field profile as $E(y,w)=\sum_m a_m(w)\phi_m(y)$, the intensity contains cross terms,
\begin{equation}
  |E|^2 =
  \sum_m |a_m\phi_m|^2
  + 2\mathrm{Re}\sum_{m<n} a_m a_n^* \phi_m\phi_n^* .
\label{eq:interference}
\end{equation}
Thus a modal scan must reintroduce cross-mode interaction before decoding.
We apply a lightweight residual MLP along the explicit mode axis,
\begin{equation}
  \tilde U_w^{+}
  =
  \tilde U_w+\mathcal{G}_{\theta}(\mathrm{LN}(\tilde U_w)),
  \qquad
  Z^{+}=\Pi_{\theta}\!\left(\{\tilde U_w^{+}\}_{w=1}^{W}\right),
\label{eq:coupling}
\end{equation}
where $\tilde U_w=[\tilde{\mathbf{u}}_{1,w},\ldots,\tilde{\mathbf{u}}_{M,w}]$, $\mathcal{G}_{\theta}$ mixes modes at the same propagation position, and $\Pi_{\theta}$ projects coupled modal tokens back to the spatial feature map $Z^+$.

For the PaNO-R2 variant, we add a lightweight output-aware feedback path in parallel with the forward backbone.
It targets output-side discontinuities and weak reflected or high-frequency residual components that are less exposed to a purely forward traveling-wave update.
R2 revisits $X^0$ in a reverse axial order as a reciprocity-inspired sensitivity refinement and produces a spatial residual $R^b=\Psi_{\mathrm{R2}}(X^0)$.
The feedback path is fused in the spatial feature domain as
\begin{equation}
  R^b=\Psi_{\mathrm{R2}}(X^0),
  \qquad
  Z^{\mathrm{R2}} = Z^{+}+R^b,
  \qquad
  \hat E = \mathrm{Head}\!\left(\mathrm{Refine}(Z^{\mathrm{R2}})\right),
\label{eq:decode}
\end{equation}
where $Z^{+}$ and $R^b$ have matching spatial resolution and channels.
R2 is therefore reported as a residual variant of PaNO: it tests whether output-side feedback can correct field components that are weakly represented by a purely forward propagation backbone.

\section{Experiments}
\label{sec:experiments}

\subsection{Experimental Setup}

We evaluate tunable $3{\times}3$ MMI waveguides as a representative and compute-feasible testbed for propagation-dominated photonic design with localized port readouts.
MMI devices combine multimode interference, self-imaging, output-side discontinuities, and port-window power aggregation, making them a compact setting for the same field-to-readout mismatch that appears in many waveguide splitters and couplers.
The benchmark uses $80\times384$ grids, a fixed test split, and a deterministic diagnostic script, with 15 wavelength settings from $1.530$ to $1.565\,\mu$m and three input-port excitations per geometry group; after flattening the port-excitation dimension, the held-out test split contains $4608$ complex-field prediction cases.
NeurOLight is the primary baseline; FNO, FactorFNO, and UNet are broader neural field baselines \citep{li2021fno,tran2023ffno,ronneberger2015unet,gu2022neurolight}.
All methods are full-field surrogates: given $\hat E$, the shared evaluator computes
\begin{equation}
\hat E
\longrightarrow
\left\{
d_F(\hat E,E),\;
d_M(\mathcal{M}(\hat E),\mathcal{M}(E)),\;
d_R(\mathcal{R}(\hat E),\mathcal{R}(E))
\right\}.
\label{eq:eval-protocol}
\end{equation}
Here $d_F$ contains dense-field errors, $d_M$ contains propagation and output-window diagnostics, and $d_R$ contains port readouts such as power, phase, coupling, and splitting.
All metrics are derived from the same $\hat E$, so improvements cannot come from a separate scalar readout head.

\subsection{Controlled Baseline Comparison}
\label{sec:main-results}

Table~\ref{tab:controlled-baselines} compares PaNO and PaNO-R2 with neural field surrogates under the same 15-wavelength MMI diagnostic protocol.

\begin{table}[H]
  \caption{Controlled comparison on the 15-wavelength tunable $3{\times}3$ MMI benchmark. All entries are computed from the predicted complex field using the same evaluation script. Lower is better.}
  \label{tab:controlled-baselines}
  \centering
  \scriptsize
  \resizebox{\columnwidth}{!}{ 
  \begin{tabular}{lcccccccc}
    \toprule
    & \multicolumn{1}{c}{Field} & \multicolumn{3}{c}{Mediator} & \multicolumn{4}{c}{Readout} \\
    \cmidrule(lr){2-2}\cmidrule(lr){3-5}\cmidrule(lr){6-9}
    Model & cMAE$\downarrow$ & SWR$\downarrow$ & Prop.\ profile$\downarrow$ & Output profile$\downarrow$ & Port power$\downarrow$ & Rel.\ phase$\downarrow$ & Coupling$\downarrow$ & Split$\downarrow$ \\
    \midrule
    FNO(2D) & 0.2107 & 0.6575 & 0.0809 & 0.5965 & 0.1531 & 0.7762 & 0.0178 & 0.0527 \\
    FactorFNO(2D) & 0.1752 & 0.0589 & 0.0459 & 0.1453 & 0.0815 & 0.7821 & 0.0177 & 0.0424 \\
    UNet & 0.8525 & 3.1010 & 0.5058 & 0.9995 & 0.9992 & 1.5315 & 0.0296 & 0.3517 \\
    NeurOLight & 0.1750 & 0.1491 & 0.1119 & 0.3646 & 0.2018 & \textbf{0.6061} & \textbf{0.0119} & 0.0429 \\
    PaNO & 0.1822 & 0.0659 & 0.0492 & 0.1732 & 0.0739 & 0.8187 & 0.0181 & 0.0396 \\
    PaNO-R2 & \textbf{0.1471} & \textbf{0.0499} & \textbf{0.0356} & \textbf{0.1001} & \textbf{0.0551} & 0.7090 & 0.0150 & \textbf{0.0324} \\
    \bottomrule
  \end{tabular}
  }
\end{table}

The completed rows show that PaNO trades slightly worse cMAE than NeurOLight for better propagation-profile, output-profile, and port-power errors, while PaNO-R2 gives the best cMAE, SWR, propagation profile, output profile, port power, and splitting.
Here output profile compares the transverse intensity distribution at the final propagation plane before port aggregation, whereas port power integrates intensity inside fixed localized port masks.
Phase and coupling are more coherence-sensitive, so the mixed results mark the boundary of our claim: PaNO primarily improves propagation-mediated power readouts rather than all port-level quantities.

In terms of efficiency, PaNO-R2 keeps the same millisecond-scale inference regime as NeurOLight: one forward pass takes $6.19$ ms on an NVIDIA RTX 5090 GPU.
The logged Angler/FDFD reference-generation pipeline takes $14$--$16$ s per instance under the same dataset workflow, so neural inference is roughly three orders of magnitude faster than generating a new reference field.

\begin{figure}[H]
  \centering
  \includegraphics[width=0.98\columnwidth]{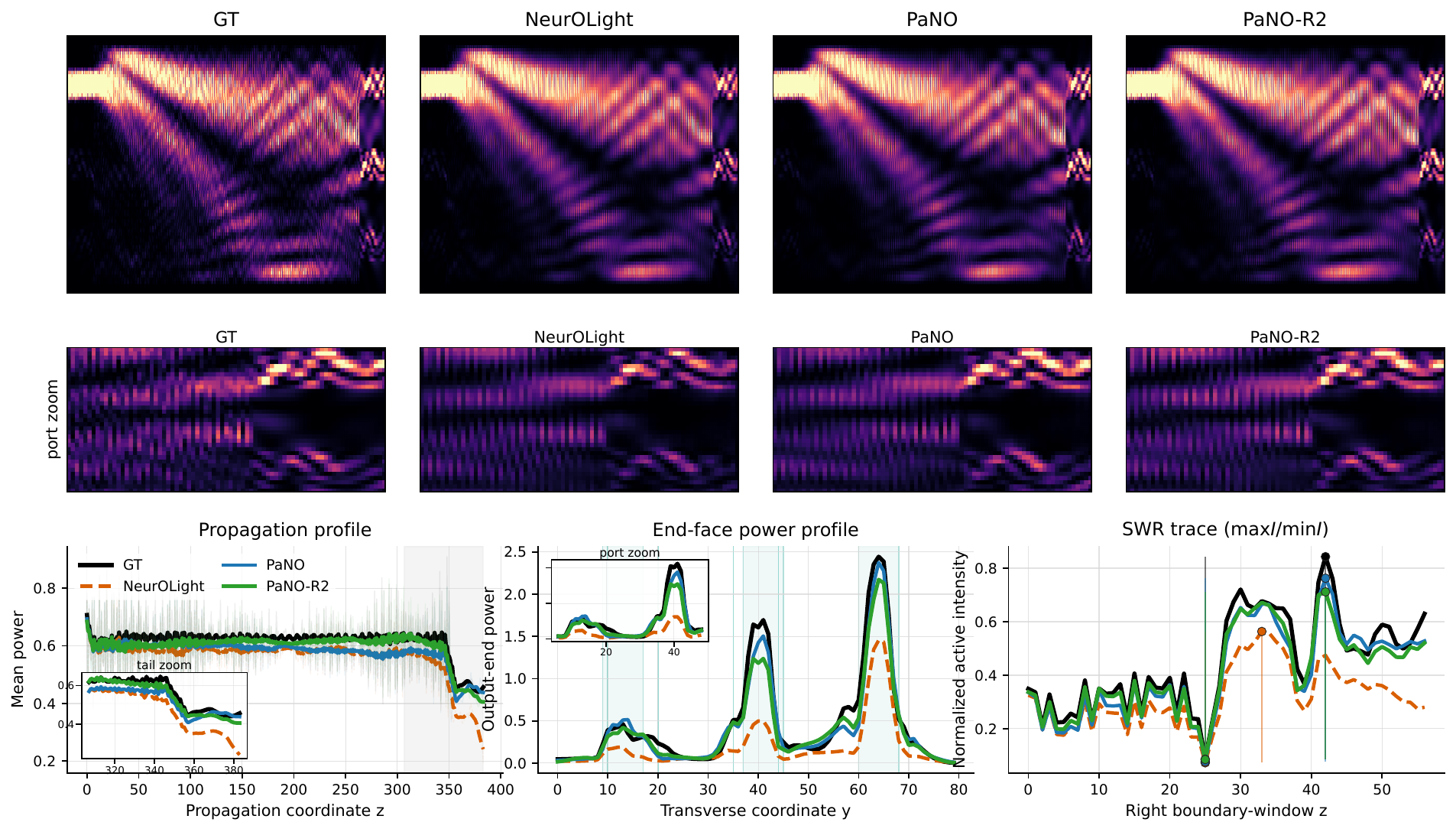}
  \caption{Qualitative interpretation of the controlled comparison. Columns show ground truth, NeurOLight, PaNO, and PaNO-R2; rows show the global field, output-side crop, propagation profile, output-plane power profile, and SWR trace. The panels illustrate how similar-looking fields can differ in propagation mediators and localized port readouts.}
  \label{fig:main-qualitative-comparison}
\end{figure}

PaNO instead propagates modal states causally from left to right, which better follows the longitudinal power envelope; PaNO-R2 further improves the output-side profile and the right-boundary SWR trace.
The visual pattern is consistent with Table~\ref{tab:controlled-baselines}: gains appear not only in cMAE for PaNO-R2, but also in propagation profile, SWR, output profile, and port-power readout.
This qualitative case is not used as standalone evidence; it motivates the aggregate diagnostic analysis below, where we test whether field-level error and output-profile error predict readout failure across the full test split.

\subsection{Diagnostic Analysis: From Field Error to Readout Failure}
\label{sec:diagnostic-analysis}

We next test the diagnostic premise from Section~\ref{sec:problem}: active-region cMAE should be reported, but localized port readout also depends on propagation and output-profile mediators.
As a quantitative check, Appendix~\ref{app:two-bound-derivation} compares an active-cMAE-based error bound with a local output-profile error bound on the same $4608$ held-out fields.
The active-cMAE-based bound is $135$--$183{\times}$ looser in median gap, supporting output-profile mediators when diagnosing port readout.

\begin{figure}[t]
  \centering
  \begin{minipage}[t]{0.495\linewidth}
    \centering
    \includegraphics[width=\linewidth]{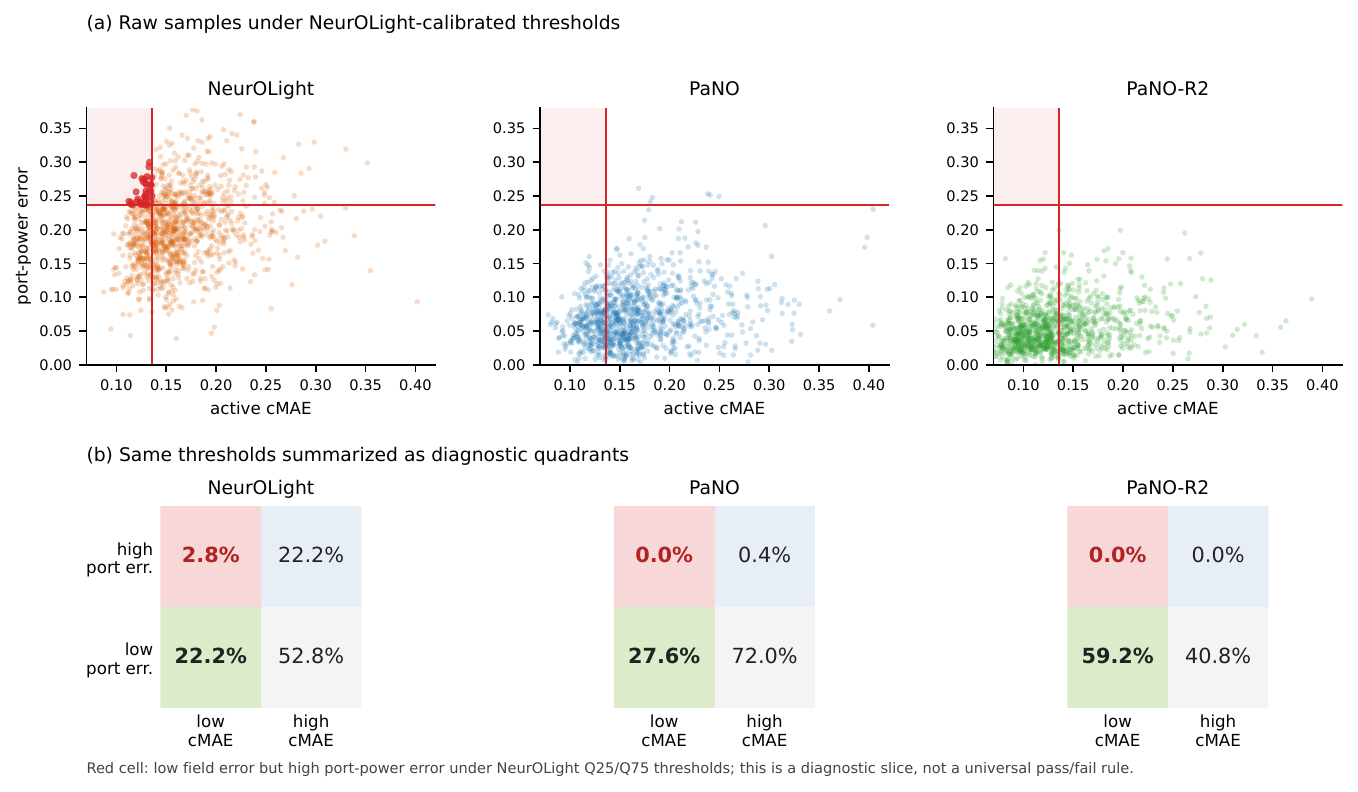}
  \end{minipage}\hfill
  \begin{minipage}[t]{0.495\linewidth}
    \centering
    \includegraphics[width=\linewidth]{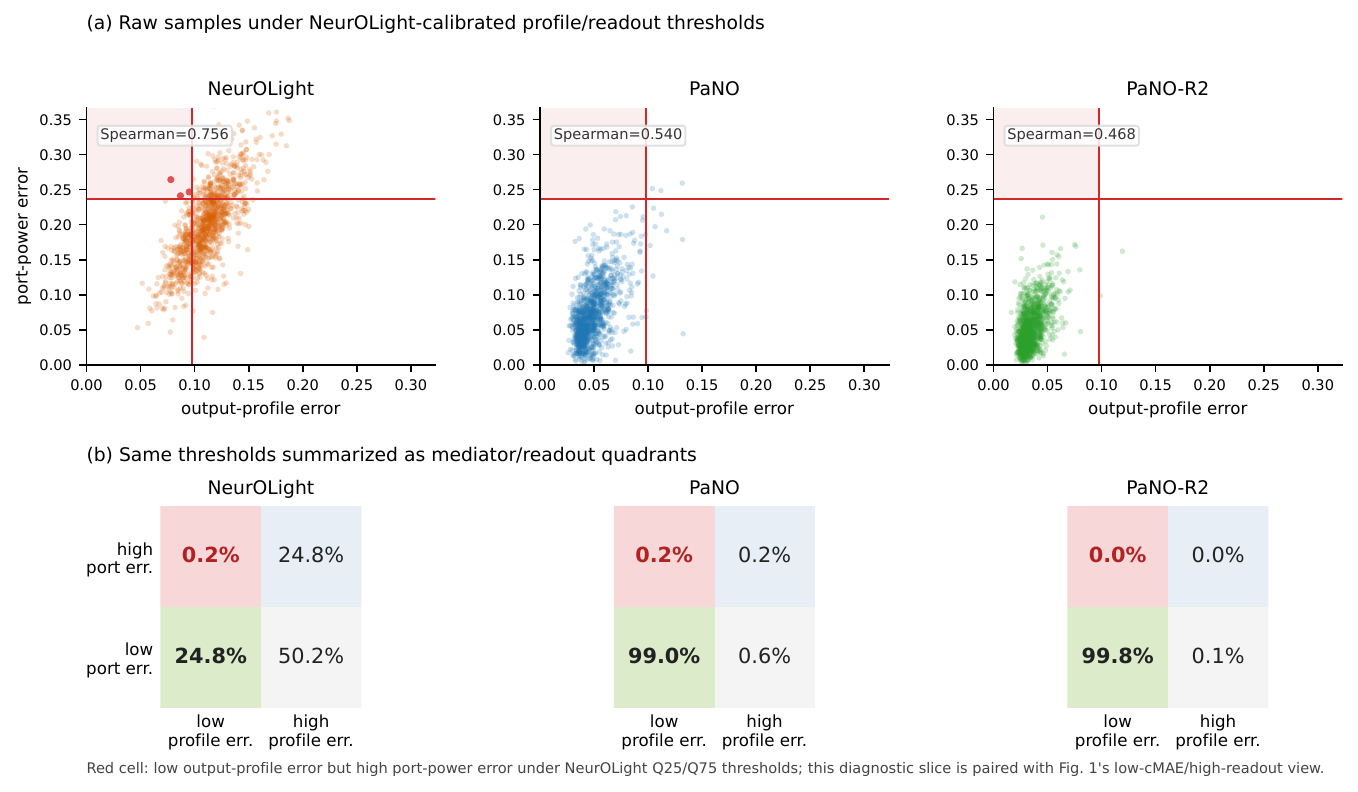}
  \end{minipage}
  \vspace{-0.4em}
  \caption{Diagnostic comparison between a dense field proxy and a readout-aligned mediator. The red quadrant marks low proxy error but high port-power error. Active-region cMAE leaves many failures in this region, whereas output-profile error aligns more directly with port-power error because it reflects the output-plane distribution before port integration.}
  \label{fig:diagnostic-panels}
\end{figure}

Figure~\ref{fig:diagnostic-panels}, left, asks whether active-region cMAE ranks port failures within each fixed model.
Low active-region cMAE can still coincide with large port-power error, and the Spearman correlations are only $0.277$, $0.210$, and $0.251$ for NeurOLight, PaNO, and PaNO-R2.
This is a sample-level statement, not a cross-model rejection of cMAE: Table~\ref{tab:controlled-baselines} shows PaNO-R2 improves cMAE and port power together, while PaNO improves port power despite slightly worse cMAE.

Equation~\eqref{eq:port-power} points to the local output profile as the relevant mediator, because port power is a localized output-window intensity integral.
Figure~\ref{fig:diagnostic-panels}, right, confirms this ranking shift: output-profile error raises the Spearman association with port-power error to $0.756$, $0.540$, and $0.468$.
This does not claim that the output profile explains phase or coupling.
It shows that local propagation quantities better diagnose port-power failures than active-region cMAE alone.
The full correlation table is in Appendix~\ref{app:diagnostics}.

Finally, the dense-field loss gives limited direct pressure to ports: the output-port union covers $2.87\%$ of the grid and contributes only $4.5$--$4.9\%$ of the residual mass.
PaNO nevertheless has a larger ratio of parameter-gradient norms computed from port-region versus full-grid residuals than NeurOLight ($0.225$ vs.\ $0.141$), consistent with a representation that keeps propagation-to-output information active.
We treat this as sensitivity evidence.
Module-level causality is tested in the ablations below.
\FloatBarrier

\subsection{Ablation Studies}
\label{sec:ablation-arch}

We use matched 100-epoch ablations on the same 15-wavelength MMI protocol to separate cross-mode coupling, local front-end bias, R2 compensation, and task-aligned fine-tuning.
These within-ablation trends are not replacements for the 200-epoch main comparison.

\begin{table}[H]
  \caption{Forward architecture ablation on the 15-wavelength MMI protocol. Lower is better.}
  \label{tab:ablation-arch}
  \centering
  \small
  \resizebox{\columnwidth}{!}{
  \begin{tabular}{lcccccc}
    \toprule
    Model & Active cMAE$\downarrow$ & Active phase$\downarrow$ & SWR$\downarrow$ & Tail/head$\downarrow$ & Output profile$\downarrow$ & Port power$\downarrow$ \\
    \midrule
    NeurOLight & 0.1857 & 0.1747 & 0.1428 & 2.5533 & 0.3329 & 0.1355 \\
    PaNO & 0.1698 & 0.1618 & 0.0655 & \textbf{1.8627} & \textbf{0.1708} & 0.0729 \\
    w/o coupling & 0.1991 & 0.1924 & 0.1276 & 2.4677 & 0.2785 & 0.1246 \\
    pointwise/window stem & \textbf{0.1284} & \textbf{0.1168} & \textbf{0.0645} & 2.3919 & 0.1823 & \textbf{0.0637} \\
    \bottomrule
  \end{tabular}
  }
\end{table}

Table~\ref{tab:ablation-arch} shows that removing cross-mode coupling worsens active-region error, SWR, output profile, and port power.
The stem ablation is a trade-off rather than a clean win for MSAS: the pointwise/window stem fits this in-distribution 100-epoch setting better on active cMAE, SWR, and port power, while PaNO retains better tail/head stability and output-profile error.
The parameter counts are nearly matched, so we interpret this as an inductive-bias trade-off rather than a capacity effect.
Our rationale for keeping MSAS is architectural: it delays strong modal mixing until the explicit modal-token and cross-mode-coupling stages, reducing the risk that early convolutional mixing contaminates the separable modal structure used by the propagation backbone.
Thus MSAS is the propagation-aligned default, but not a universally dominant front end.

\begin{table}[H]
  \caption{Readout-sensitive variants on the 15-wavelength MMI protocol. R2 changes the architecture, while stage-2 port fine-tuning changes the objective. Lower is better.}
  \label{tab:ablation-r2-task}
  \centering
  \small
  \resizebox{\columnwidth}{!}{
  \begin{tabular}{lccccccc}
    \toprule
    Variant & Active cMAE$\downarrow$ & SWR$\downarrow$ & Tail/head$\downarrow$ & Output profile$\downarrow$ & Port power$\downarrow$ & Rel.\ phase$\downarrow$ & Coupling$\downarrow$ \\
    \midrule
    PaNO & 0.1698 & 0.0655 & \textbf{1.8627} & 0.1708 & 0.0729 & 0.8056 & 0.0177 \\
    PaNO-R2 & \textbf{0.1614} & \textbf{0.0552} & 1.9209 & \textbf{0.1228} & 0.0684 & 0.6775 & \textbf{0.0139} \\
    stage-2 port FT & 0.1694 & 0.0700 & 1.9644 & 0.1590 & \textbf{0.0581} & \textbf{0.5391} & 0.0796 \\
    \bottomrule
  \end{tabular}
  }
\end{table}

Table~\ref{tab:ablation-r2-task} separates two readout interventions: R2 improves dense-field, SWR, output profile, and coupling with a tail-stability cost, while port fine-tuning gives the best port power and relative phase but severely worsens coupling.
Task-aligned supervision is therefore a deployment knob that moves the readout Pareto point, not a substitute for a propagation-aligned field model.

\subsection{Generalization and Target-Domain Fine-Tuning}
\label{sec:generalization}

We evaluate target-domain adaptation on two same-topology shifted MMI settings.
The wavelength-transfer task adapts to target wavelengths held out from the source training split, testing whether a model can adjust to changed propagation phase and effective modal constants without changing the device topology.
The refractive-index task adapts to shifted material-index conditions in the same nominal $3{\times}3$ device family, testing whether the model can recalibrate propagation constants, interference patterns, and output-window power profiles under material perturbations.
For each target setting, we initialize from the corresponding source checkpoint, freeze the backbone and train the output layers for 20 epochs (Linear Probing, LP), then unfreeze the full model and fine-tune all parameters for 30 epochs (FT).

\begin{table}[H]
  \caption{Target-domain fine-tuning results on same-topology shifted MMI settings. All entries are computed after LP20+FT30 adaptation from the predicted complex field; lower is better.}
  \label{tab:generalization-ft}
  \centering
  \scriptsize
  \resizebox{\columnwidth}{!}{
  \begin{tabular}{llcccccc}
    \toprule
    & & Field & Mediator & \multicolumn{4}{c}{Readout} \\
    \cmidrule(lr){3-3}\cmidrule(lr){4-4}\cmidrule(lr){5-8}
    Target shift & Model & cMAE$\downarrow$ & SWR$\downarrow$ & Port power$\downarrow$ & Rel.\ phase$\downarrow$ & Coupling$\downarrow$ & Split$\downarrow$ \\
    \midrule
    \multirow{3}{*}{Wavelength transfer}
      & NeurOLight & 0.2741 & 0.4188 & 0.2395 & 0.8419 & 0.0211 & 0.0517 \\
      & PaNO & 0.2884 & 0.1233 & 0.1673 & 1.0807 & 0.0288 & 0.0631 \\
      & PaNO-R2 & \textbf{0.2280} & \textbf{0.0701} & \textbf{0.1165} & \textbf{0.7741} & \textbf{0.0181} & \textbf{0.0447} \\
    \midrule
    \multirow{3}{*}{Refractive-index shift}
      & NeurOLight & 0.2227 & 0.1484 & 0.1734 & 0.8650 & \textbf{0.0172} & 0.0581 \\
      & PaNO & 0.2099 & 0.0782 & \textbf{0.1138} & 0.9277 & 0.0211 & 0.0503 \\
      & PaNO-R2 & \textbf{0.1818} & \textbf{0.0750} & 0.1169 & \textbf{0.8420} & 0.0178 & \textbf{0.0458} \\
    \bottomrule
  \end{tabular}
  }
\end{table}

Table~\ref{tab:generalization-ft} shows that PaNO-R2 is strongest on all reported field, mediator, and readout metrics for wavelength transfer.
This is the setting where the device topology, output-port layout, and coupling graph remain fixed; the main change is a continuous shift in propagation phase and effective modal constants.
The learned modal basis, axial scan, and cross-mode coupling therefore remain reusable after target-domain fine-tuning, while the R2 branch can correct output-side residuals that directly affect port readout.
For refractive-index shift, PaNO-R2 gives the best cMAE, SWR, relative phase, and splitting, while PaNO is slightly better on port-power error.
This pattern is consistent with a structured adaptation view: material perturbations change propagation constants and local interference statistics, but LP20+FT30 can recalibrate the modal dynamics and coupling layers with limited target supervision.
Thus the same-topology transfer results support our main claim that propagation-aligned parameterizations are especially useful when target-domain fine-tuning can adjust continuous physical parameters without relearning a new port topology.

\section{Conclusion}
We studied full-field neural surrogates for photonic devices whose design objectives are localized port readouts.
The central conclusion is that dense field accuracy remains necessary, but it is not a complete design proxy without propagation and output-window diagnostics.
Accordingly, we introduced a Field/Mediator/Readout evaluation view and PaNO, a propagation-aligned neural operator that preserves transverse modal organization and directed propagation within a full-field surrogate.

On the 15-wavelength MMI benchmark, PaNO improves propagation-profile, output-profile, and port-power errors despite slightly worse cMAE than NeurOLight, while PaNO-R2 achieves the best cMAE and port-power error among completed main rows.
The evidence is limited to frequency-domain 2D Hz MMI devices with fixed localized ports, and phase-sensitive readouts remain challenging.
Extending this propagation-readout perspective to broader photonic components and vectorial simulations is future work.

\bibliographystyle{plainnat}
\bibliography{metric_device_refs}

\clearpage
\appendix
\setcounter{page}{1}
\renewcommand{\thepage}{A-\arabic{page}}
\numberwithin{figure}{section}
\numberwithin{table}{section}
\numberwithin{equation}{section}
\makeatletter
\renewcommand{\@seccntformat}[1]{%
  \@ifundefined{#1@cntformat}%
  {\csname the#1\endcsname\quad}%
  {\csname #1@cntformat\endcsname}}
\@namedef{section@cntformat}{Appendix \thesection:\quad}
\makeatother

\section{Dataset and Physical Setting}
\label{app:dataset}

Our main empirical setting is a tunable $3{\times}3$ MMI waveguide benchmark derived from frequency-domain simulations of Hz-polarized optical fields.
Each example asks a model to predict the complex steady-state field on an $80\times384$ grid from the relative-permittivity layout, the input-port excitation, the wavelength, and grid-step information.
The ``15-wavelength'' protocol used in the main text combines 15 wavelength settings from $1.530$ to $1.565\,\mu$m at $2.5$ nm spacing.
For each wavelength setting, the processed dataset contains 512 geometry/condition groups, and each group contains three input-port excitations; this gives 7680 groups, or 23040 complex-field instances after flattening the excitation dimension.
The deterministic split used by all completed main-table models contains 6144 training groups and 1536 held-out test groups, corresponding to 18432 training and 4608 test field instances after flattening.

The benchmark is physically useful because MMI devices are dominated by multimode propagation and coherent self-imaging: small changes in wavelength, geometry, input port, or output-end phase can redistribute energy across output ports.
This makes the dataset a natural stress test for the paper's central claim.
A model must reconstruct the full complex field, but the quantities that matter for device use are localized port powers, splitting behavior, phase relations, and coupling-like readouts.
The grid spacing varies with the generated device instance, with values in the processed split spanning approximately $0.0677$--$0.0938\,\mu$m; the resulting simulated window covers the MMI propagation region at subwavelength resolution.
The relative-permittivity values span the low-index background/cladding and high-index guiding regions, with the processed range bounded by approximately $1.0$ and $12.3$.
We use the historical directory name \texttt{lyy\_raw\_tunable\_15pol} only in code and experiment manifests; in the paper, we refer to this setting as the 15-wavelength tunable $3{\times}3$ MMI benchmark because the indexed \texttt{rHz} files correspond to Hz-polarized wavelength-condition groups rather than distinct physical polarizations.

\section{From Helmholtz Equation to Modal Stepping}
\label{app:modal-derivation}

This appendix gives the derivation behind Eq.~\eqref{eq:modal-stepping}.
For a 2D Hz-polarized field in a nonmagnetic medium, the frequency-domain scalar equation can be written abstractly as
\begin{equation}
\mathcal{L}_{\omega,\varepsilon}u=s,
\qquad
\mathcal{L}_{\omega,\varepsilon}
=
\nabla\!\cdot\!\alpha(y,w)\nabla + k_0^2,
\qquad
\alpha(y,w)=\varepsilon_r(y,w)^{-1},
\label{eq:app-hz-helmholtz}
\end{equation}
up to the source and boundary-condition conventions used by the FDFD solver.
After discretization on the Yee/FDFD grid, Eq.~\eqref{eq:app-hz-helmholtz} gives the linear system $A_{\omega,\varepsilon}u=b$ used in Eq.~\eqref{eq:helmholtz-fdfd}.

To expose the propagation structure, split the coordinates into transverse position $y$ and propagation position $w$.
At a fixed cross section $w$, define transverse modes $\phi_m(y;w)$ by the local eigenproblem
\begin{equation}
\mathcal{L}_{\perp,w}\phi_m(y;w)=\beta_m^2(w)\rho_w(y)\phi_m(y;w),
\qquad
\langle \phi_m,\phi_n\rangle_{\rho_w}=\delta_{mn},
\label{eq:app-transverse-eigen}
\end{equation}
where $\mathcal{L}_{\perp,w}$ is the transverse part of the Helmholtz operator, $\beta_m(w)$ is the local propagation constant, $\rho_w$ is the weight induced by the discretization, and $\langle\cdot,\cdot\rangle_{\rho_w}$ is the corresponding weighted inner product.
Expanding the field in this moving basis,
\begin{equation}
u(y,w)=\sum_{m=1}^{M} a_m(w)\phi_m(y;w)+u_{\perp}(y,w),
\label{eq:app-modal-expansion}
\end{equation}
separates the retained modal coefficients $\mathbf a(w)=[a_1(w),\ldots,a_M(w)]^\top$ from the truncation residual $u_{\perp}$.
Substituting Eq.~\eqref{eq:app-modal-expansion} into Eq.~\eqref{eq:app-hz-helmholtz} and projecting with $\langle \phi_n,\cdot\rangle_{\rho_w}$ yields a coupled modal system
\begin{equation}
\frac{d^2\mathbf a}{dw^2}
+ C_1(w)\frac{d\mathbf a}{dw}
+ C_0(w)\mathbf a
=
\mathbf f(w)+\boldsymbol{\tau}(w),
\label{eq:app-second-order-modal}
\end{equation}
where $C_0$ contains the local propagation constants and material terms, $C_1$ contains basis-variation coupling from $\partial_w\phi_m$, $\mathbf f$ is the projected source, and $\boldsymbol{\tau}$ collects truncation, boundary, and unmodeled radiation terms.
In a uniform section, the coupling matrices vanish and Eq.~\eqref{eq:app-second-order-modal} reduces to independent modal waves $a_m(w)\propto e^{\pm i\beta_m w}$.

Classical EME turns this second-order modal equation into a section-wise transfer problem.
For a short section $[w_j,w_{j+1}]$ with nearly constant geometry, the retained forward and backward modal amplitudes are propagated by
\begin{equation}
\mathbf z_{j+1}
=
\underbrace{
S_{j+1}^{-1}S_j
\begin{bmatrix}
\mathrm{diag}(e^{i\boldsymbol{\beta}_j\Delta w}) & 0\\
0 & \mathrm{diag}(e^{-i\boldsymbol{\beta}_j\Delta w})
\end{bmatrix}}_{\mathcal{T}^{\mathrm{EME}}_j}
\mathbf z_j
+\mathbf q_j,
\label{eq:app-eme-transfer}
\end{equation}
where $\mathbf z_j$ stacks modal amplitudes at section $j$, $S_j$ maps between adjacent local modal bases, $\Delta w=w_{j+1}-w_j$, and $\mathbf q_j$ accounts for sources, boundary effects, truncation, and radiation not represented by the retained modes.
If the device is dominated by left-to-right propagation, the backward components can be absorbed into a residual term, giving the first-order state form
\begin{equation}
\frac{d\mathbf a}{dw}=K_{\omega,\varepsilon(w)}\mathbf a(w)+\mathbf r(w),
\qquad
\mathbf a_{w+1}\approx T_{\omega,\varepsilon(w)}\mathbf a_w+\mathbf q_w,
\label{eq:app-first-order-stepping}
\end{equation}
which is the physical counterpart of Eq.~\eqref{eq:modal-stepping}.
Here $K_{\omega,\varepsilon(w)}$ is the continuous generator of modal evolution, $T_{\omega,\varepsilon(w)}$ is its discrete section transfer, and $\mathbf r,\mathbf q$ collect reflected waves, output discontinuities, basis truncation, sources, and boundary corrections.
PaNO uses this state-update organization as an inductive bias: its learned tokens and SSM states are not physical EME coefficients, but they follow the same decomposition into transverse modal content, propagation-axis state transport, and residual correction.

\section{Metric Definitions}
\label{app:metrics}

Let $\Omega$ denote the full spatial grid, $\Omega_{\mathrm{act}}$ an active region extracted from the ground-truth field magnitude, $\Omega_{\mathrm{int}}$ a material-interface band, and $\mathcal{P}$ the set of output ports.
We use a small $\epsilon>0$ only to avoid division by zero in normalized ratios.
All metrics are lower-is-better unless explicitly stated.

\paragraph{Dense-field cMAE.}
\begin{equation}
\mathrm{cMAE} =
\frac{\sum_{x\in\Omega}|E_{\mathrm{pred}}(x)-E_{\mathrm{gt}}(x)|}
{\sum_{x\in\Omega}|E_{\mathrm{gt}}(x)|+\epsilon}.
\end{equation}

\paragraph{Active-region cMAE.}
\begin{equation}
\mathrm{cMAE}_{\mathrm{act}} =
\frac{\sum_{x\in\Omega_{\mathrm{act}}}|E_{\mathrm{pred}}(x)-E_{\mathrm{gt}}(x)|}
{\sum_{x\in\Omega_{\mathrm{act}}}|E_{\mathrm{gt}}(x)|+\epsilon}.
\end{equation}

\paragraph{Active-region phase error.}
\begin{equation}
\mathrm{PhaseMAE}_{\mathrm{act}} =
\frac{1}{|\Omega_{\mathrm{act}}|}
\sum_{x\in\Omega_{\mathrm{act}}}
\left|\mathrm{wrap}\!\left(\phi_{\mathrm{pred}}(x)-\phi_{\mathrm{gt}}(x)\right)\right|.
\end{equation}

\paragraph{Propagation and output diagnostics.}
Let $I_E(x)=|E(x)|^2$ and let $w$ index the propagation axis.
The axial propagation profile used in the main table is
\begin{equation}
g_E(w)=\frac{1}{|\Omega_w|}\sum_{x\in\Omega_w} I_E(x),
\qquad
\mathrm{PropProfileErr}=
\frac{\|g_{E_{\mathrm{pred}}}-g_{E_{\mathrm{gt}}}\|_1}
{\|g_{E_{\mathrm{gt}}}\|_1+\epsilon},
\end{equation}
where $\Omega_w$ is the transverse slice at position $w$.
Output-profile error applies the same normalized $L_1$ comparison to the final-plane transverse intensity profile,
\begin{equation}
\mathrm{OutputProfileErr}=
\frac{\|I_{E_{\mathrm{pred}}}(\cdot,W)-I_{E_{\mathrm{gt}}}(\cdot,W)\|_1}
{\|I_{E_{\mathrm{gt}}}(\cdot,W)\|_1+\epsilon}.
\end{equation}
For a fixed boundary-sensitive trace $b_E(w)$, standing-wave-ratio error compares the peak-valley contrast
\begin{equation}
\mathrm{SWR}(E)=
\frac{\max_w b_E(w)+\epsilon}{\min_w b_E(w)+\epsilon},
\qquad
\mathrm{SWRErr}=|\mathrm{SWR}(E_{\mathrm{pred}})-\mathrm{SWR}(E_{\mathrm{gt}})|.
\end{equation}
Tail/head error measures output-side error amplification relative to input-side error:
\begin{equation}
\mathrm{TailHeadErr}=
\frac{\mathrm{MAE}(E_{\mathrm{pred}},E_{\mathrm{gt}};\Omega_{\mathrm{tail}})+\epsilon}
{\mathrm{MAE}(E_{\mathrm{pred}},E_{\mathrm{gt}};\Omega_{\mathrm{head}})+\epsilon}.
\end{equation}
The regions $\Omega_w$, $\Omega_{\mathrm{tail}}$, $\Omega_{\mathrm{head}}$, and the boundary trace $b_E$ are fixed by the shared deterministic evaluator for all models.

\paragraph{Port readout diagnostics.}
For each output port $p\in\mathcal{P}$, the port power is
\begin{equation}
P_p(E)=\sum_{x\in\Omega_p}m_p(x)|E(x)|^2,
\qquad
\mathrm{PortPowerErr}=
\frac{1}{|\mathcal{P}|}\sum_{p\in\mathcal{P}}
\frac{|P_p(E_{\mathrm{pred}})-P_p(E_{\mathrm{gt}})|}
{P_p(E_{\mathrm{gt}})+\epsilon}.
\end{equation}
Relative phase error computes wrapped phase differences between port pairs,
\begin{equation}
\mathrm{RelPhaseErr}=
\frac{1}{|\mathcal{Q}|}\sum_{(p,q)\in\mathcal{Q}}
\left|\mathrm{wrap}\!\left(
[\phi_p(E_{\mathrm{pred}})-\phi_q(E_{\mathrm{pred}})]
-[\phi_p(E_{\mathrm{gt}})-\phi_q(E_{\mathrm{gt}})]
\right)\right|,
\end{equation}
where $\phi_p(E)$ is the evaluator's port-window phase summary and $\mathcal{Q}$ is the fixed set of port pairs.
Coupling error compares the evaluator-defined complex coupling vector $c(E)$,
\begin{equation}
\mathrm{CouplingErr}=
\frac{\|c(E_{\mathrm{pred}})-c(E_{\mathrm{gt}})\|_1}
{\|c(E_{\mathrm{gt}})\|_1+\epsilon}.
\end{equation}
Splitting error compares normalized port-power distributions
\begin{equation}
s_p(E)=\frac{P_p(E)}{\sum_{q\in\mathcal{P}}P_q(E)+\epsilon},
\qquad
\mathrm{SplitErr}=\frac{1}{|\mathcal{P}|}\sum_{p\in\mathcal{P}}
|s_p(E_{\mathrm{pred}})-s_p(E_{\mathrm{gt}})|.
\end{equation}

\section{Training Details}
\label{app:training}

The main 15-wavelength MMI comparison and the ablation studies use different training budgets, so their absolute numbers should not be mixed.
The main comparison uses archived checkpoints from 200-epoch training runs for all reported baseline and PaNO-family rows.
Ablations use a separate 100-epoch budget to compare variants under matched ablation cost; these results are interpreted as within-ablation trends rather than direct replacements for the 200-epoch main comparison.
All models are evaluated using the same field-output interface, the same metric scripts, and fixed checkpoint paths archived in the experiment manifest.
The controlled comparison table omits parameter counts because its purpose is the Field/Mediator/Readout diagnostic comparison; efficiency and model-size considerations are separated from readout correctness.

\begin{table}[H]
  \caption{Protocol summary for reported comparison rows. The main table uses the same 15-wavelength MMI test split and deterministic evaluation script for all entries; training hyperparameters are listed to make remaining differences explicit.}
  \label{tab:app-protocol}
  \centering
  \scriptsize
  \setlength{\tabcolsep}{4pt}
  \begin{tabularx}{\linewidth}{@{}l c X c X@{}}
    \toprule
    Model group & Epochs & Objective & Effective batch & Role \\
    \midrule
    FNO & 200 & dense-field NMAE & 4 & broader baseline \\
    FactorFNO & 200 & dense-field NMAE & 4 & broader baseline \\
    UNet & 200 & dense-field NMAE & 4 & broader baseline \\
    NeurOLight & 200 & dense-field NMAE & 4 & primary baseline \\
    PaNO & 200 & dense-field cMAE & 4 & ours, forward model \\
    PaNO-R2 & 200 & dense-field cMAE & 4 & ours, R2 variant \\
    Ablation variants & 100 & dense-field cMAE or marked task loss & 4 & within-ablation only \\
    \bottomrule
  \end{tabularx}
\end{table}

\begin{table}[H]
  \caption{Checkpoint selection for the controlled baseline table. ``Retained'' means the checkpoint kept by the corresponding training run's checkpoint-selection rule; all rows are then re-evaluated by the same full15 diagnostic script.}
  \label{tab:app-checkpoints}
  \centering
  \scriptsize
  \setlength{\tabcolsep}{4pt}
  \begin{tabularx}{\linewidth}{@{}l c X@{}}
    \toprule
    Model & Reported checkpoint epoch & Selection note \\
    \midrule
    FNO & 200 & retained New\_Model checkpoint \\
    FactorFNO & 194 & retained New\_Model checkpoint \\
    UNet & 195 & retained New\_Model checkpoint \\
    NeurOLight & 200 & archived main baseline checkpoint \\
    PaNO & 200 & archived main PaNO checkpoint \\
    PaNO-R2 & 200 & archived main PaNO-R2 checkpoint \\
    \bottomrule
  \end{tabularx}
\end{table}

\begin{table}[H]
  \caption{Compute time for the reported main-comparison and ablation runs. All rows use one RTX 5090 worker; times are reported as pure runtime and exclude idle gaps between resumed segments.}
  \label{tab:app-compute-main}
  \centering
  \scriptsize
  \setlength{\tabcolsep}{4pt}
  \begin{tabularx}{\linewidth}{@{}X r r@{}}
    \toprule
    Run & Params & Pure runtime (h) \\
    \midrule
    FNO & 3.2872M & 2.526 \\
    FactorFNO & 3.1629M & 11.409 \\
    UNet & 3.4658M & 2.304 \\
    NeurOLight & 1.5888M & 5.503 \\
    PaNO & 2.7020M & 25.513 \\
    PaNO-R2 & 3.0519M & 27.156 \\
    w/o coupling ablation & 2.6976M & 10.887 \\
    stage-2 port FT & 2.7020M & 4.387 \\
    pointwise/window stem ablation & 2.6848M & 9.636 \\
    \midrule
    Reported subtotal & -- & 98.501 \\
    \bottomrule
  \end{tabularx}
\end{table}

\section{Statistical Stability}
\label{app:stability}

The main tables use a fixed deterministic train/test split and one archived checkpoint per model row, while Table~\ref{tab:app-seed-stability} adds a two-seed retraining check for the principal NeurOLight, PaNO, and PaNO-R2 comparison.
All seed rows are evaluated by the same deterministic Field/Mediator/Readout script on the same $4608$ held-out complex-field cases, so the reported variation reflects initialization and training-order effects under a fixed benchmark rather than changes in the test set.
The result supports the qualitative ranking in the main table: PaNO variants remain stronger on propagation-mediated and port-power readouts, while NeurOLight remains competitive on phase-sensitive quantities.

\begin{table}[H]
  \caption{Two-seed retraining stability on the 15-wavelength MMI benchmark. Entries are mean $\pm$ standard deviation over two independently trained checkpoints, evaluated on the same held-out test split. Lower is better.}
  \label{tab:app-seed-stability}
  \centering
  \scriptsize
  \resizebox{\columnwidth}{!}{
  \begin{tabular}{lcccccc}
    \toprule
    Model & cMAE & SWR & Prop.\ profile & Output profile & Port power & Rel.\ phase \\
    \midrule
    NeurOLight & $0.1762 \pm 0.0076$ & $0.1571 \pm 0.0610$ & $0.0978 \pm 0.0191$ & $0.3736 \pm 0.0651$ & $0.1819 \pm 0.0478$ & $0.6919 \pm 0.1464$ \\
    PaNO & $\mathbf{0.1279 \pm 0.0122}$ & $0.0574 \pm 0.0092$ & $\mathbf{0.0394 \pm 0.0007}$ & $0.1601 \pm 0.0360$ & $\mathbf{0.0584 \pm 0.0009}$ & $\mathbf{0.6645 \pm 0.0432}$ \\
    PaNO-R2 & $0.1555 \pm 0.0095$ & $\mathbf{0.0517 \pm 0.0003}$ & $0.0404 \pm 0.0032$ & $\mathbf{0.1162 \pm 0.0111}$ & $0.0643 \pm 0.0002$ & $0.7167 \pm 0.0758$ \\
    \bottomrule
  \end{tabular}
  }
\end{table}

The two-seed check should be read as a stability supplement rather than a replacement for the paired main-table comparison.
It shows that the lower PaNO/PaNO-R2 propagation and port-power errors are not produced by a single favorable checkpoint, but it also preserves the boundary of the claim: phase-sensitive readouts remain mixed and should not be described as uniformly improved.

\section{Additional Generalization Results}
\label{app:generalization}

This appendix expands the target-domain fine-tuning evidence beyond the two same-topology settings reported in the main text.
For every target-domain adaptation run, the model is initialized from the corresponding source-domain checkpoint, the backbone is frozen while the output layers are trained for 20 epochs (LP), and then all parameters are unfrozen for 30 epochs of full fine-tuning (FT).
The tables below report only the final LP20+FT30 endpoint.
They should not be pooled with the controlled 15-wavelength in-distribution benchmark: each transfer setting has its own target-domain dataset, and the historical FFT/stem suite additionally uses archived 100-epoch checkpoints and partially different metric families.

\subsection{Current Three-Model Transfer Suite}
\label{app:current-transfer-suite}

Table~\ref{tab:app-current-transfer-ft} reports the full Field/Mediator/Readout diagnostics for the current NeurOLight, PaNO, and PaNO-R2 transfer suite.
The four target settings cover port-count topology transfer, wavelength transfer, and refractive-index shift.
The two same-topology shifts support the main-text adaptation claim, while the two topology-transfer rows document the boundary case where the number of ports and output coupling graph change.

\begin{table}[H]
  \caption{Current three-model target-domain fine-tuning suite. All entries are final LP20+FT30 endpoints. Lower is better for cMAE, RelL2, SWR, port power, and relative phase; higher is better for F1.}
  \label{tab:app-current-transfer-ft}
  \centering
  \scriptsize
  \resizebox{\columnwidth}{!}{
  \begin{tabular}{llrrrrrr}
    \toprule
    Target shift & Model & cMAE & RelL2 & F1 & SWR & Port power & Rel. phase \\
    \midrule
    \multirow{3}{*}{$3{\times}3\!\rightarrow\!4{\times}4$}
      & NeurOLight & \textbf{0.2718} & \textbf{0.0805} & \textbf{0.9595} & \textbf{0.2796} & \textbf{0.2278} & \textbf{0.9279} \\
      & PaNO & 0.3150 & 0.1057 & 0.9437 & 0.7124 & 0.2995 & 1.2306 \\
      & PaNO-R2 & 0.3066 & 0.1003 & 0.9505 & 0.3239 & 0.2298 & 1.1239 \\
    \midrule
    \multirow{3}{*}{Wavelength transfer}
      & NeurOLight & 0.2741 & 0.0924 & 0.9674 & 0.4188 & 0.2395 & 0.8419 \\
      & PaNO & 0.2884 & 0.0914 & 0.9587 & 0.1233 & 0.1673 & 1.0807 \\
      & PaNO-R2 & \textbf{0.2280} & \textbf{0.0579} & \textbf{0.9681} & \textbf{0.0701} & \textbf{0.1165} & \textbf{0.7741} \\
    \midrule
    \multirow{3}{*}{$3{\times}3\!\rightarrow\!5{\times}5$}
      & NeurOLight & 0.4559 & 0.2158 & \textbf{0.9173} & \textbf{0.5972} & \textbf{0.4515} & \textbf{1.1304} \\
      & PaNO & \textbf{0.4319} & \textbf{0.1924} & 0.9085 & 1.0619 & 0.4731 & 1.3387 \\
      & PaNO-R2 & 0.4787 & 0.2386 & 0.9147 & 0.8408 & 0.4548 & 1.2923 \\
    \midrule
    \multirow{3}{*}{Refractive-index shift}
      & NeurOLight & 0.2227 & 0.0498 & 0.9615 & 0.1484 & 0.1734 & 0.8650 \\
      & PaNO & 0.2099 & 0.0424 & 0.9569 & 0.0782 & \textbf{0.1138} & 0.9277 \\
      & PaNO-R2 & \textbf{0.1818} & \textbf{0.0329} & \textbf{0.9641} & \textbf{0.0750} & 0.1169 & \textbf{0.8420} \\
    \bottomrule
  \end{tabular}
  }
\end{table}

The current transfer suite shows a structured pattern rather than uniform dominance.
For wavelength transfer, PaNO-R2 is best on every reported field, mediator, and readout metric; this is the cleanest same-topology adaptation result.
For refractive-index shift, PaNO-R2 is best on cMAE, RelL2, F1, SWR, and relative phase, while PaNO is slightly better on port-power error.
For topology transfer, the conclusion is deliberately weaker.
In $3{\times}3\!\rightarrow\!4{\times}4$, NeurOLight is strongest across all reported metrics after fine-tuning.
In $3{\times}3\!\rightarrow\!5{\times}5$, PaNO gives the lowest cMAE and RelL2, but NeurOLight remains better on F1, SWR, port-power error, and relative-phase error.
This supports the interpretation that changing the port topology is a harder structural transfer problem than shifting wavelength or refractive index within the same topology.

The refractive-index rows in Table~\ref{tab:app-current-transfer-ft} use the corrected evaluation normalization for the shifted-index target domain.
The corrected evaluator uses the same permittivity range as the fine-tuning run rather than reloading the checkpoint under the default dataset range.
This matters because an inconsistent permittivity normalization can make the same trained checkpoint appear to fail under the refractive-index evaluator.

\subsection{Historical FFT/Stem Transfer Suite}
\label{app:historical-fft-stem}

Table~\ref{tab:app-historical-compact-ft} summarizes the historical NeurOLight-100e, NeuroMamba-FFT, and NeuroMamba-Stem transfer endpoints.
These models are not the current PaNO/PaNO-R2 checkpoints.
They are included to document model evolution: the early FFT/reference line and later pointwise-window-stem line already showed target-domain adaptation capacity in several shifted settings, but the evidence is metric-specific and should not replace the current controlled comparison.

\begin{table}[H]
  \caption{Historical target-domain fine-tuning endpoints for archived 100-epoch source models. Rows report only the final LP20+FT30 result. The wavelength row uses cMAE/RelL2 from the newer diagnostic evaluator; the other rows follow the historical N-MAE/RelL2 reports.}
  \label{tab:app-historical-compact-ft}
  \centering
  \scriptsize
  \resizebox{\columnwidth}{!}{
  \begin{tabular}{llrrr}
    \toprule
    Target shift & Metric & NeurOLight-100e & NeuroMamba-FFT & NeuroMamba-Stem \\
    \midrule
    $3{\times}3\!\rightarrow\!4{\times}4$ & N-MAE & 0.289962 & \textbf{0.242325} & 0.268117 \\
    $3{\times}3\!\rightarrow\!4{\times}4$ & RelL2 & 0.092835 & 0.268979 & \textbf{0.080213} \\
    Wavelength transfer & cMAE & \textbf{0.257301} & 0.277029 & 0.328806 \\
    Wavelength transfer & RelL2 & \textbf{0.070460} & 0.099644 & 0.113486 \\
    $3{\times}3\!\rightarrow\!5{\times}5$ & N-MAE & 0.189558 & 0.140436 & \textbf{0.138008} \\
    $3{\times}3\!\rightarrow\!5{\times}5$ & RelL2 & 0.035538 & 0.026031 & \textbf{0.019314} \\
    Refractive-index shift & N-MAE & 0.261260 & \textbf{0.166610} & 0.178931 \\
    Refractive-index shift & RelL2 & 0.067554 & 0.033048 & \textbf{0.031610} \\
    \bottomrule
  \end{tabular}
  }
\end{table}

The compact historical table shows why these runs are useful but not clean enough to be the main paper evidence.
For $3{\times}3\!\rightarrow\!4{\times}4$, FFT gives the best final N-MAE but Stem gives the best final RelL2.
For wavelength transfer, the historical NeurOLight-100e row remains better than the two historical NeuroMamba rows on cMAE and RelL2.
For $3{\times}3\!\rightarrow\!5{\times}5$ and refractive-index shift, the historical NeuroMamba variants are stronger than NeurOLight on the final field-error endpoints.
Thus the historical suite supports an adaptation-capacity and model-evolution story, not a uniform dominance claim.

\subsection{Historical Wavelength Diagnostics}
\label{app:historical-wavelength-diagnostics}

The historical wavelength-transfer run is the only historical setting with a full paper-quality diagnostic table for all three archived models.
Table~\ref{tab:app-historical-wavelength-ft} reports the final fine-tuned endpoint only.
It is useful because it separates field accuracy, support quality, phase, SWR, port power, and relative phase under the same diagnostic evaluator.

\begin{table}[H]
  \caption{Historical wavelength-transfer diagnostics after LP20+FT30. These are archived 100-epoch source models, not the current PaNO/PaNO-R2 checkpoints. Lower is better except for F1.}
  \label{tab:app-historical-wavelength-ft}
  \centering
  \scriptsize
  \resizebox{\columnwidth}{!}{
  \begin{tabular}{lrrrrrrrr}
    \toprule
    Model & cMAE & RelL2 & cMAE$_{\mathrm{act}}$ & F1 & Phase$_{\mathrm{act}}$ & SWR & Port power & Rel. phase \\
    \midrule
    NeurOLight-100e & \textbf{0.257301} & \textbf{0.070460} & \textbf{0.245016} & 0.960369 & \textbf{0.231960} & 0.301131 & \textbf{0.210132} & \textbf{0.818519} \\
    NeuroMamba-FFT & 0.277029 & 0.099644 & 0.266999 & \textbf{0.965042} & 0.261048 & 0.428690 & 0.323530 & 1.330838 \\
    NeuroMamba-Stem & 0.328806 & 0.113486 & 0.315122 & 0.950607 & 0.318870 & \textbf{0.150505} & 0.218178 & 0.937870 \\
    \bottomrule
  \end{tabular}
  }
\end{table}

This diagnostic table explains why the historical wavelength-transfer setting is not used as a positive NeuroMamba-family claim.
NeurOLight-100e is stronger on most final field and readout errors, while NeuroMamba-Stem is best only on SWR and NeuroMamba-FFT is best only on support F1.
The current PaNO-R2 wavelength-transfer result in Table~\ref{tab:app-current-transfer-ft} is therefore the relevant evidence for the current model, not the older FFT/stem wavelength-transfer result.

\subsection{Unified FFT Re-Evaluation}
\label{app:fft-unified-reeval}

The archived NeuroMamba-FFT checkpoint was also re-evaluated with the newer paper-quality script across all four shifted settings.
These rows are diagnostic continuity checks for one historical model.
They should not be substituted for Table~\ref{tab:app-historical-compact-ft}, because the newer evaluator and the historical N-MAE/RelL2 reports are not identical metric families.

\begin{table}[H]
  \caption{Unified newer-evaluator re-check of the historical NeuroMamba-FFT fine-tuned checkpoints. Rows report final adapted endpoints only.}
  \label{tab:app-fft-unified-ft}
  \centering
  \scriptsize
  \resizebox{\columnwidth}{!}{
  \begin{tabular}{lrrrr}
    \toprule
    Target shift & cMAE & RelL2 & cMAE$_{\mathrm{act}}$ & Port power \\
    \midrule
    $3{\times}3\!\rightarrow\!4{\times}4$ & 0.378543 & 0.160939 & 0.364468 & 0.362896 \\
    Wavelength transfer & 0.277029 & 0.099644 & 0.266999 & 0.323530 \\
    $3{\times}3\!\rightarrow\!5{\times}5$ & 0.382902 & 0.151402 & 0.369725 & 0.180196 \\
    Refractive-index shift & 1.251527 & 1.793230 & 1.252342 & 0.302570 \\
    \bottomrule
  \end{tabular}
  }
\end{table}

The unified FFT re-evaluation is mixed.
It gives a reasonable port-power endpoint for the $3{\times}3\!\rightarrow\!5{\times}5$ transfer, but the refractive-index row is poor under this newer diagnostic evaluator.
This reinforces the control-variable caution above: historical rows are useful for audit and model evolution, while the current NeurOLight/PaNO/PaNO-R2 tables are the appropriate basis for claims about the final model family.

\subsection{Two-Bound Validation}
\label{app:two-bound-derivation}
\label{app:diagnostics}

This subsection makes explicit the two conservative bounds used in the diagnostic discussion.
The goal is not to introduce a deep theorem, but to clarify why a dense field error can be valid yet weak for localized port-power diagnosis, while an output-profile error is more tightly aligned with the readout.

Let $u\in\mathbb{C}^{H\times W}$ be the ground-truth field, $\hat u$ the predicted field, and $e=\hat u-u$ the field residual.
For a port mask $m_p(x)\in[0,1]$ supported on $\Omega_p$, the evaluator uses the discrete intensity readout
\begin{equation}
P_p(u)=\sum_{x\in\Omega_p}m_p(x)|u(x)|^2,
\qquad
\Delta P_p=P_p(\hat u)-P_p(u).
\label{eq:app-port-readout}
\end{equation}
Write $M_p=\max_{x\in\Omega_p}m_p(x)$ and $\|v\|_{1,\Lambda}=\sum_{x\in\Lambda}|v(x)|$ for any region $\Lambda$.
The first identity is
\begin{equation}
\Delta P_p
=
\sum_{x\in\Omega_p}m_p(x)\Big(|u(x)+e(x)|^2-|u(x)|^2\Big)
=
\sum_{x\in\Omega_p}m_p(x)\Big(2\Re(u^*(x)e(x))+|e(x)|^2\Big).
\label{eq:app-port-expansion}
\end{equation}
Using $||a+b|^2-|a|^2|\le 2|a||b|+|b|^2\le (2|a|+|b|)|b|$ gives
\begin{equation}
|\Delta P_p|
\le
\sum_{x\in\Omega_p}m_p(x)(2|u(x)|+|e(x)|)|e(x)|.
\label{eq:app-port-basic-bound}
\end{equation}

\paragraph{Bound A: dense-field cMAE bound.}
Let
\begin{equation}
\bar e_{\Omega}=\frac{1}{|\Omega|}\|e\|_{1,\Omega},
\qquad
\bar e_{\Omega_p}=\frac{1}{|\Omega_p|}\|e\|_{1,\Omega_p},
\qquad
\eta_p=\frac{|\Omega_p|}{|\Omega|}.
\label{eq:app-avg-errors}
\end{equation}
If the residual is moderate relative to the field magnitude, Eq.~\eqref{eq:app-port-basic-bound} yields the linearized bound
\begin{equation}
|\Delta P_p|
\le
2M_p\|u\|_{1,\Omega_p}\bar e_{\Omega_p}
\le
2M_p\|u\|_{1,\Omega}\bar e_{\Omega_p}.
\label{eq:app-local-cmae-bound}
\end{equation}
To express this in terms of a dense global average, one must convert the unknown local mean error $\bar e_{\Omega_p}$ to the observed global mean error $\bar e_{\Omega}$.
In the worst case,
\begin{equation}
\bar e_{\Omega_p}\le \frac{1}{\eta_p}\bar e_{\Omega},
\label{eq:app-area-ratio}
\end{equation}
because the same total $L_1$ error can be concentrated entirely inside the small port region.
Combining Eqs.~\eqref{eq:app-local-cmae-bound} and \eqref{eq:app-area-ratio} gives
\begin{equation}
|\Delta P_p|
\le
\frac{2M_p\|u\|_{1,\Omega}}{\eta_p}\bar e_{\Omega}.
\label{eq:app-global-cmae-bound}
\end{equation}
Equation~\eqref{eq:app-global-cmae-bound} is legally correct, but its looseness is driven by the area ratio $1/\eta_p$.
For localized output masks, $\eta_p\ll 1$, so a small dense-field cMAE does not prevent large port error if the residual mass is concentrated near the output plane.

The same logic applies if the evaluator reports active-region cMAE rather than full-grid cMAE.
Replacing $\Omega$ by $\Omega_{\mathrm{act}}$ shrinks the averaging domain, but the bound still inherits an area-ratio factor
\begin{equation}
\eta_p^{\mathrm{act}}=\frac{|\Omega_p|}{|\Omega_{\mathrm{act}}|},
\qquad
\bar e_{\Omega_p}\le \frac{1}{\eta_p^{\mathrm{act}}}\bar e_{\Omega_{\mathrm{act}}},
\label{eq:app-active-area-ratio}
\end{equation}
so it remains a valid but potentially loose certificate.

\paragraph{Bound B: output-profile bound.}
Define the final-plane transverse intensity profile
\begin{equation}
q_u(y)=|u(y,W)|^2,
\qquad
q_{\hat u}(y)=|\hat u(y,W)|^2,
\label{eq:app-output-profile}
\end{equation}
and let the output-profile error be
\begin{equation}
B_p
=
\sum_{y\in\Omega_p}m_p(y)\big|q_{\hat u}(y)-q_u(y)\big|.
\label{eq:app-output-profile-bound}
\end{equation}
Because the port readout itself is a masked sum over this same final-plane intensity,
\begin{equation}
|\Delta P_p|
=
\left|
\sum_{y\in\Omega_p}m_p(y)\big(q_{\hat u}(y)-q_u(y)\big)
\right|
\le
B_p.
\label{eq:app-readout-aligned-bound}
\end{equation}
This bound is readout-aligned: it uses exactly the quantity that is later aggregated by the port operator, with no domain-size conversion from the full field to a small output window.
If the signed intensity error keeps a consistent sign inside the port, the bound is nearly tight; if positive and negative fluctuations cancel, the gap is at most a small constant-factor effect from absolute-value removal rather than the large area-ratio factor in Bound A.

\paragraph{Optimization view.}
The same mismatch appears in the training loss.
For a dense $L_1$-style field objective
\begin{equation}
\mathcal{L}_{\mathrm{field}}
=
\frac{1}{|\Omega|}\sum_{x\in\Omega}|e(x)|,
\label{eq:app-dense-loss}
\end{equation}
the pointwise gradient magnitude with respect to the prediction is approximately uniform:
\begin{equation}
\left|\frac{\partial \mathcal{L}_{\mathrm{field}}}{\partial \hat u(x)}\right|
\propto
\frac{1}{|\Omega|}.
\label{eq:app-dense-grad}
\end{equation}
Therefore the total direct gradient budget assigned to a port region scales only with its area fraction,
\begin{equation}
\sum_{x\in\Omega_p}
\left|
\frac{\partial \mathcal{L}_{\mathrm{field}}}{\partial \hat u(x)}
\right|
=
O(\eta_p).
\label{eq:app-port-grad-budget}
\end{equation}
This does not mean dense-field supervision is wrong; it means that localized readout quality can only improve reliably if the representation transports upstream propagation information into the small output region.
This is the motivation for the Field/Mediator/Readout diagnostics and for PaNO's propagation-structured latent states.

Empirically, the same $4608$ held-out fields show that the active-cMAE-based bound is $135$--$183{\times}$ looser in median gap than the output-profile bound.
That numerical gap is exactly the behavior predicted by Eqs.~\eqref{eq:app-global-cmae-bound}--\eqref{eq:app-readout-aligned-bound}: both bounds are valid, but only the output-profile bound is tightly matched to the localized port readout.

The main text reports the strongest diagnostic evidence: weak active-cMAE/readout rank correlation and a port-power formula linking readout error to output-profile error.
As a secondary diagnostic, we also train a lightweight random-forest predictor with five-fold cross-validation to predict readout-error ranks from Field features, Mediator features, or both.
Field features include cMAE, relative $L_2$, and active-region cMAE; Mediator features include standing-wave contrast error, tail/head error amplification, output-profile error, and edge-profile error.
Field+Mediator improves over Field alone for port power, relative phase, coupling, and splitting on average, with Spearman correlations of 0.527, 0.178, 0.339, and 0.199, respectively.
The active-cMAE/readout Spearman correlations are weak across the completed main models: for NeurOLight, PaNO, and PaNO-R2, the correlations with port power are 0.250, 0.226, and 0.217; with relative phase, 0.119, 0.127, and 0.120; with coupling, 0.180, 0.219, and 0.254; and with splitting, 0.181, 0.169, and 0.171.

For the port-power readout emphasized in the main text, propagation/output-profile errors have stronger rank association with port-power error than active cMAE: 0.756/0.540/0.468 for NeurOLight/PaNO/PaNO-R2, compared with 0.277/0.210/0.251 for active cMAE in the same analysis.
The full analysis also counts samples that fall in the best quartile by active cMAE but the worst quartile by a readout metric.
For NeurOLight, this low-cMAE/high-readout-error mismatch appears in 125, 245, 206, and 179 samples for port power, relative phase, coupling, and splitting, respectively.
For PaNO, the corresponding counts are 164, 205, 171, and 203; for PaNO-R2, they are 159, 247, 149, and 200.
These counts are not used to claim that one model is categorically more failure-prone than another, because the quartiles are computed within each model.
Their purpose is to show that low global field error and poor endpoint behavior can coexist in a non-negligible subset of cases.
The main text instead uses unified NeurOLight thresholds for model comparison, which separates the definition of the danger region from the model being evaluated.

The qualitative panels in Appendix~\ref{app:qualitative} use this diagnostic selection principle.
They show representative output-sensitive field crops and profile strips rather than isolated scalar readout bars, so the visual evidence remains tied to the same mediator quantities used in the main quantitative analysis.
The examples are illustrative only; the main claims are based on the aggregate tables and the fixed-split diagnostic statistics above.

\section{Additional Ablations}
\label{app:additional-ablations}

This appendix supplements the ablation section with a Transformer backbone control.

\paragraph{Transformer backbone control.}
NeuroTransformer1D keeps the full-field surrogate interface and uses a parameter-matched Transformer encoder along the propagation axis.
The evaluated checkpoint is the retained best-validation checkpoint from the 100-epoch parameter-matched run.
Table~\ref{tab:app-transformer-backbone} reports the resulting full15 diagnostic metrics on the 4608-case held-out test split.

\begin{table}[H]
  \caption{Transformer backbone control on the 15-wavelength tunable $3{\times}3$ MMI benchmark. The NeuroTransformer1D row uses the retained best-validation checkpoint from the 100-epoch parameter-matched Transformer run; all entries are recomputed from predicted complex fields on the same 4608-case test split. Lower is better.}
  \label{tab:app-transformer-backbone}
  \centering
  \scriptsize
  \resizebox{\columnwidth}{!}{
  \begin{tabular}{lcccccccc}
    \toprule
    & \multicolumn{1}{c}{Field} & \multicolumn{3}{c}{Mediator} & \multicolumn{4}{c}{Readout} \\
    \cmidrule(lr){2-2}\cmidrule(lr){3-5}\cmidrule(lr){6-9}
    Model & cMAE$\downarrow$ & SWR$\downarrow$ & Prop.\ profile$\downarrow$ & Output profile$\downarrow$ & Port power$\downarrow$ & Rel.\ phase$\downarrow$ & Coupling$\downarrow$ & Split$\downarrow$ \\
    \midrule
    NeurOLight & 0.1750 & 0.1491 & 0.1119 & 0.3646 & 0.2018 & \textbf{0.6061} & \textbf{0.0119} & 0.0429 \\
    PaNO & 0.1822 & 0.0659 & 0.0492 & 0.1732 & 0.0739 & 0.8187 & 0.0181 & 0.0396 \\
    PaNO-R2 & \textbf{0.1471} & \textbf{0.0499} & \textbf{0.0356} & \textbf{0.1001} & \textbf{0.0551} & 0.7090 & 0.0150 & \textbf{0.0324} \\
    NeuroTransformer1D & 0.1783 & 0.1375 & 0.0739 & 0.2610 & 0.1309 & 1.0941 & 0.0295 & 0.0865 \\
    \bottomrule
  \end{tabular}}
\end{table}

The full evaluator report also records $N=4608$, $\mathrm{rel}\,L_2=0.0417$, support F1 $=0.9757$, and active phase error $=0.1577$.

\paragraph{A--F small-subset study.}
The historical A--F series tests the design axes that were later compressed into the final method: stem and modal basis choices, early reverse/backward paths, task-aligned losses, MSAS/PPDS stems, and port-loss/backward interactions.
Table~\ref{tab:app-af-inventory} gives the mapping from experiment family to evidence files.
The most useful lesson is not that every physics-inspired variant helps; rather, the early sweeps show that adaptive modal representations, light cross-mode coupling, and task-aligned losses are useful only under the right structural constraints.

\begin{table}[H]
  \caption{Inventory of the small-subset A--F ablation records. These runs are development evidence, not the main benchmark protocol.}
  \label{tab:app-af-inventory}
  \centering
  \scriptsize
  \resizebox{\columnwidth}{!}{
  \begin{tabular}{lll}
    \toprule
    Family & Main design question & Primary archived record \\
    \midrule
    A & front-end stem and FFT/DCT/learnable basis choices & \texttt{small\_dataset\_checkpoint\_collection\_A\_to\_G\_20260414.md} \\
    B & early DPAM/reverse-path structure & same A--G checkpoint inventory \\
    C & task-loss and curriculum variants on an early reverse model & same A--G checkpoint inventory \\
    D & MSAS/PPDS stems and basis combinations & \texttt{A\_to\_G\_all\_20260415.../summary\_metrics.csv} \\
    E & transferring task strategies to the D3 backbone & \texttt{A\_to\_G\_all\_20260417.../summary\_metrics.csv} \\
    F & backward-path correction combined with port loss & \texttt{fseries\_100e\_result\_report\_20260413.md} \\
    Cross-mode & sparse residual coupling between modal tokens & \texttt{crossmode\_all\_compare\_test\_full\_20260415/summary\_metrics.csv} \\
    \bottomrule
  \end{tabular}}
\end{table}

Table~\ref{tab:app-small-key} reports representative rows from the small-subset experiments.
The D-series comparison shows why the final model did not simply adopt an arbitrary physics-inspired front end: MSAS with a learnable modal basis is substantially better than MSAS+DCT or PPDS+DCT on SWR, port power, and output-edge profile error.
The E/F rows show the task-loss trade-off: port-aware losses can strongly reduce port-power error, but they often degrade dense-field or coupling-sensitive quantities.

\begin{table}[H]
  \caption{Representative small-subset ablation rows. Lower is better. These values come from the archived one-wavelength evaluator with 309 held-out field instances.}
  \label{tab:app-small-key}
  \centering
  \scriptsize
  \resizebox{\columnwidth}{!}{
  \begin{tabular}{lccccc}
    \toprule
    Model & Active cMAE & Active phase & SWR & Port power & Output profile \\
    \midrule
    NeurOLight-1wl & 0.3313 & \textbf{0.3774} & 2.4029 & 0.5212 & 0.8633 \\
    D1: MSAS+DCT & 0.3859 & 0.4567 & 3.8073 & 0.5848 & 0.9117 \\
    D2: PPDS+DCT & 0.3967 & 0.4720 & 3.2052 & 0.6031 & 0.8911 \\
    D3: MSAS+learnable & \textbf{0.3265} & 0.3839 & 1.8588 & 0.4931 & 0.8276 \\
    D4: MSAS+DCT+early reverse & 0.3847 & 0.4574 & 2.9051 & 0.5858 & 0.8887 \\
    \midrule
    E0: D3, cMAE & 0.3319 & 0.3913 & 2.1076 & 0.5221 & 0.8430 \\
    E3: balanced task curriculum & 0.4282 & 0.5033 & 1.2854 & 0.3019 & 0.7603 \\
    F1: gated reverse, cMAE & 0.3912 & 0.4708 & 1.8543 & 0.5761 & 0.8523 \\
    F2: gated reverse + port loss & 0.4314 & 0.5064 & 1.3005 & \textbf{0.2342} & \textbf{0.7600} \\
    F3: port loss, no reverse & 0.5042 & 0.6099 & 2.9175 & 0.3073 & 0.8449 \\
    \bottomrule
  \end{tabular}}
\end{table}

\paragraph{Cross-mode coupling and capacity.}
A separate small-subset sweep isolates the modal coupling module.
Table~\ref{tab:app-crossmode-small} shows that a light residual MLP across the explicit mode axis improves active-field, phase, SWR, port-power, and coupling errors over the no-coupling precursor.
Increasing coupling capacity improves some field metrics but does not monotonically improve port power, which motivated the light MLP used in the final PaNO backbone.

\begin{table}[H]
  \caption{Small-subset cross-mode coupling sweep. Lower is better.}
  \label{tab:app-crossmode-small}
  \centering
  \scriptsize
  \resizebox{\columnwidth}{!}{
  \begin{tabular}{lccccc}
    \toprule
    Model & Active cMAE & Active phase & SWR & Port power & Coupling \\
    \midrule
    no cross-mode MLP & 0.3265 & 0.3840 & 1.8607 & 0.4837 & 0.0312 \\
    light cross-mode MLP & 0.3007 & 0.3525 & 1.6744 & \textbf{0.4243} & 0.0272 \\
    MLP + attention & 0.3106 & 0.3583 & \textbf{1.5395} & 0.4654 & \textbf{0.0235} \\
    gated MLP & 0.3235 & 0.3804 & 2.4380 & 0.5016 & 0.0247 \\
    wide MLP, interval 1 & \textbf{0.2914} & \textbf{0.3345} & 1.7244 & 0.4478 & 0.0249 \\
    \bottomrule
  \end{tabular}}
\end{table}

\paragraph{Basis constraints and negative probes.}
Additional basis experiments tested orthogonality penalties and PCA/QR initialization.
They showed that basis geometry matters, but orthogonality itself is not a reliable predictor of field or readout quality: lowering $\|UU^\top-I\|_F$ did not monotonically improve active cMAE, SWR, or port power.
High-frequency residual heads and several early reverse-path variants likewise remained development diagnostics because they did not produce stable main-metric gains under later evaluation.
These negative probes are why the final paper frames the method around propagation-aligned sequence modeling and cross-mode interaction rather than around basis orthogonality or generic backward processing.

\paragraph{Long runs and compute provenance.}
The final PaNO-family rows were not selected from the small-subset suite.
They were trained on the full 15-wavelength protocol and then evaluated on the fixed 4608-instance held-out split.
On the 100-epoch full-protocol ablation, removing cross-mode coupling worsens PaNO from 0.1698 to 0.1991 active cMAE, from 0.0655 to 0.1276 SWR error, and from 0.0729 to 0.1246 port-power error, supporting cross-mode coupling as the most stable structural component.
Adding R2 improves the same PaNO checkpoint from 0.1698 to 0.1614 active cMAE, from 0.0655 to 0.0552 SWR error, and from 0.1708 to 0.1228 output-profile error, while increasing tail/head from 1.8627 to 1.9209.
Stage-2 port fine-tuning gives the best port-power and relative-phase errors, 0.0581 and 0.5391, but worsens coupling error to 0.0796; the pointwise/window stem gives the best active cMAE, 0.1284, but worse tail/head and output-profile behavior than PaNO.
The archived compute audit records PaNO 100e plus 100e-to-200e continuation as 25.513 GPU-hours, PaNO-R2 100e plus continuation as 27.156 GPU-hours, and the no-coupling, stage-2 port fine-tuning, and pointwise/window full-protocol ablations as 10.887, 4.387, and 9.636 GPU-hours, respectively, all on a single RTX 5090 worker.

\section{Additional Qualitative Visualizations}
\label{app:qualitative}

This appendix provides the remaining qualitative panels for the 200-epoch 15-wavelength MMI comparison: NeurOLight, PaNO, and PaNO-R2.
They complement the main diagnostic figures by checking whether quantitative gains correspond to output-sensitive field structure.

\begin{figure}[H]
  \centering
  \includegraphics[width=0.56\textwidth]{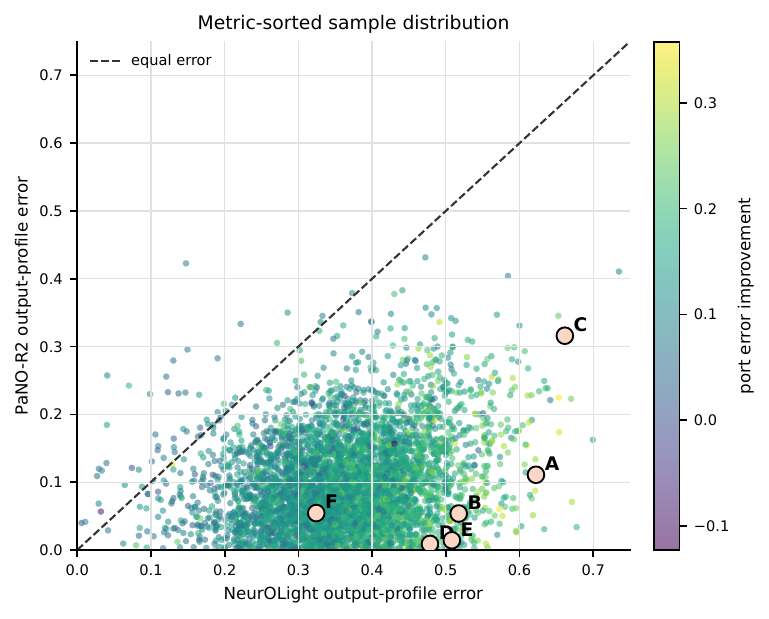}
  \vspace{-0.5em}
  \caption{Metric-sorted qualitative sample selection used for the appendix visualizations. Each point is one held-out test case, plotted by NeurOLight output-profile error (x-axis) and PaNO-R2 output-profile error (y-axis), so points below the diagonal indicate cases where PaNO-R2 improves the output-side profile that precedes port integration. The lettered samples are chosen to span several regimes: clear PaNO-R2 gains, modest changes, and cases where improvements in output-profile error do not imply uniform gains on every local feature. This panel therefore explains why the following qualitative figures focus on output-end structure rather than only on dense field appearance.}
  \label{fig:app-fig4a}
\end{figure}

\begin{figure}[H]
  \centering
  \includegraphics[width=0.52\textwidth]{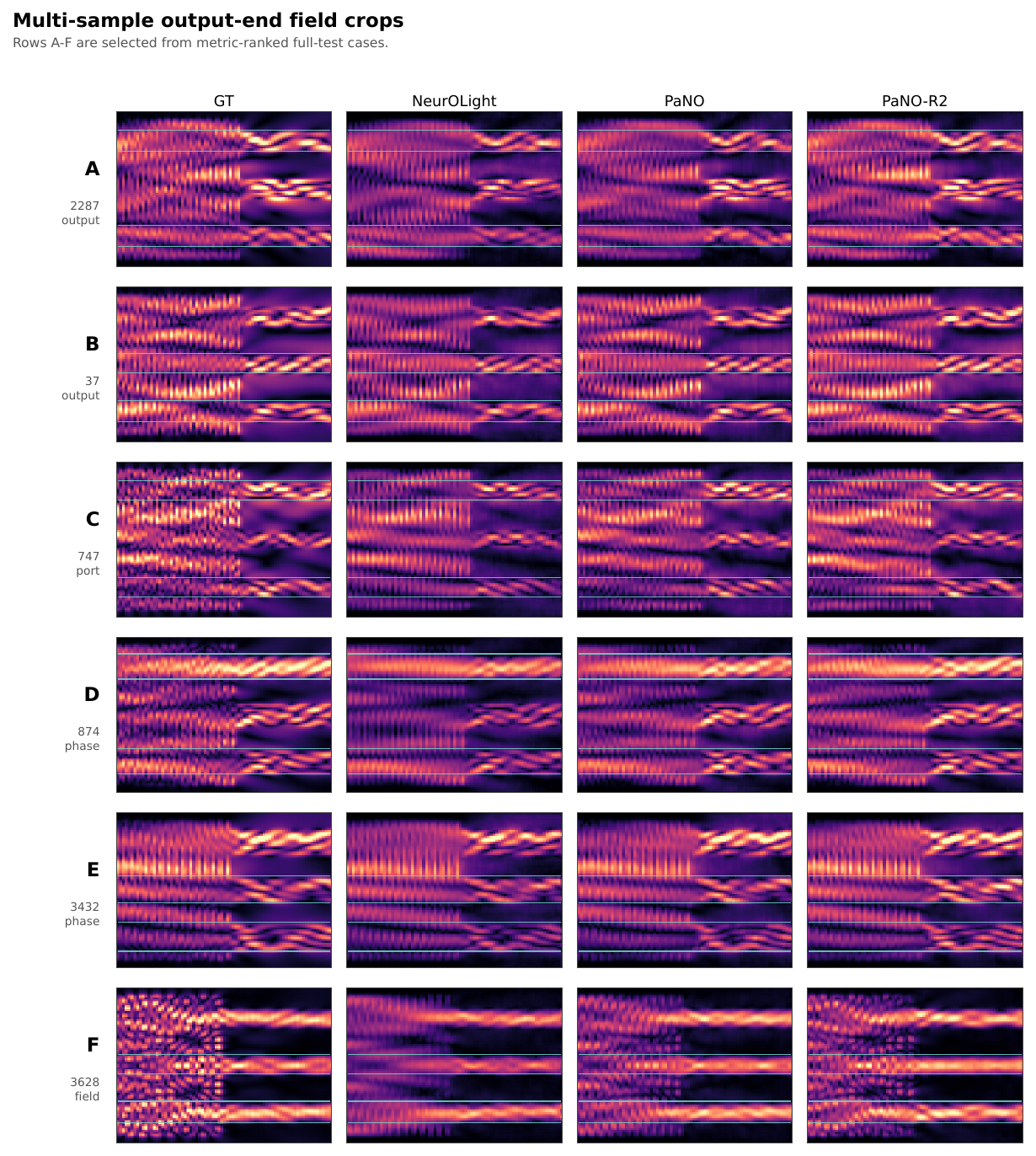}
  \vspace{-0.5em}
  \caption{Multi-sample output-end field crops for the six lettered cases selected in Figure~\ref{fig:app-fig4a}. In each row, the columns are ground truth, NeurOLight, PaNO, and PaNO-R2, shown with a shared row-wise color scale so that comparisons remain within the same physical sample. These crops isolate the output-side interference pattern that determines port-window power after integration. Across rows, NeurOLight often preserves the coarse bright-lobe layout but distorts local lobe width, sideband leakage, or dark-gap separation near the ports; PaNO usually sharpens the output structure, and PaNO-R2 most often gives the closest match to the localized output-end morphology. This is the qualitative counterpart of the main-text claim that output-profile accuracy is a better precursor of port readout than broad full-field similarity alone.}
  \label{fig:app-fig4b}
\end{figure}

\begin{figure}[H]
  \centering
  \includegraphics[width=0.58\textwidth]{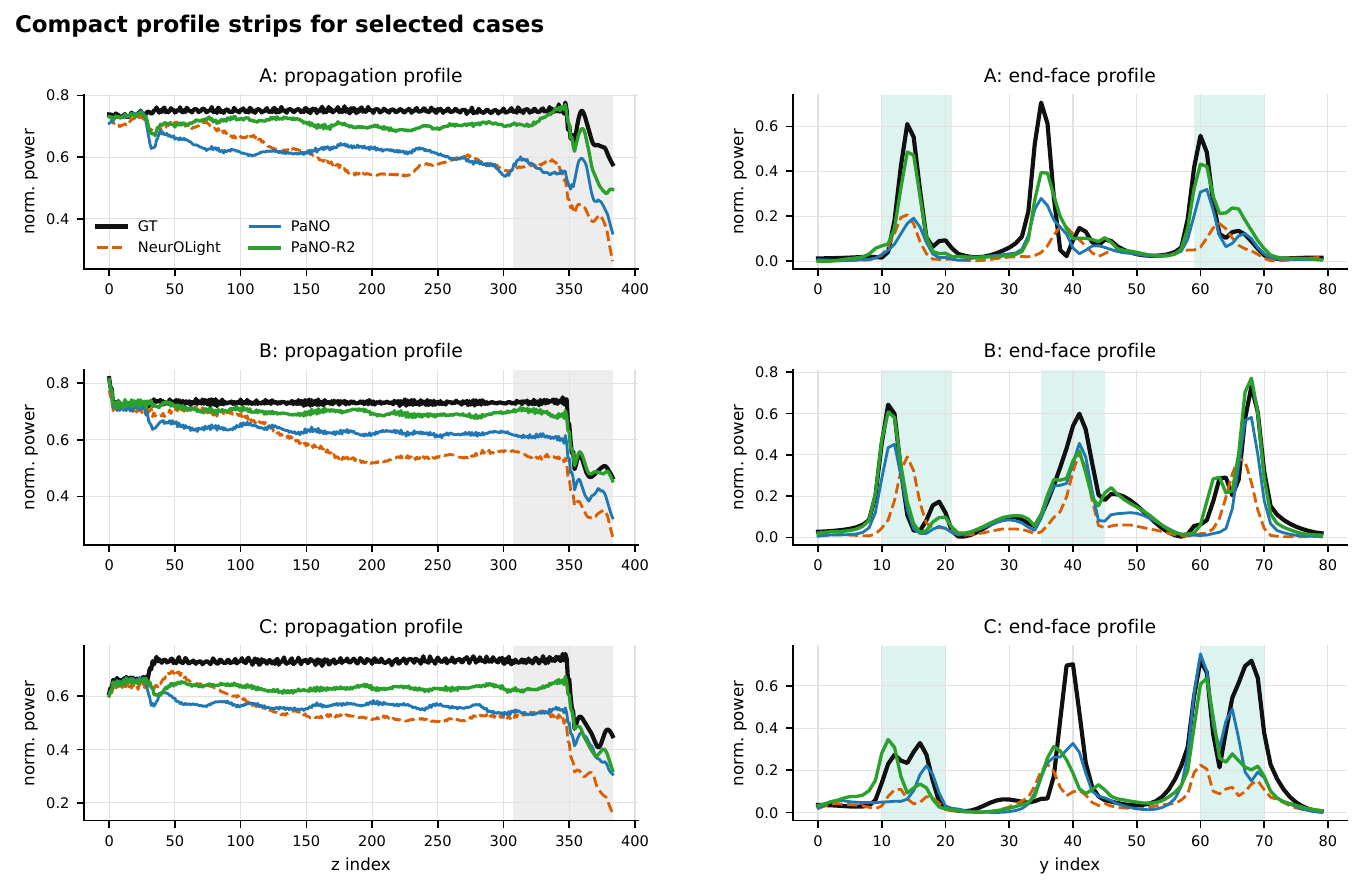}
  \vspace{-0.5em}
  \caption{Compact profile strips for three representative cases from Figure~\ref{fig:app-fig4a}. The left column compares normalized propagation power profiles along the axial coordinate, which show whether each model preserves the longitudinal transport envelope and tail-end decay before the output plane. The right column compares normalized end-face transverse power profiles across the output region, which directly reveal peak placement, lobe width, and valley separation before port integration. Read together, the two columns separate two failure modes that can look similar in a field image: a model may transport power along the device reasonably well yet still mis-shape the final transverse distribution, or it may already drift in the propagation envelope before reaching the output face. The cases here show that PaNO-R2 tends to improve both stages together more consistently than NeurOLight, while PaNO often captures the transport trend but leaves larger output-shape residuals.}
  \label{fig:app-fig4c}
\end{figure}

\clearpage
\renewcommand{\thepage}{C-\arabic{page}}
\setcounter{page}{1}
\section*{NeurIPS Paper Checklist}

\begin{enumerate}

\item {\bf Claims}
    \item[] Question: Do the main claims made in the abstract and introduction accurately reflect the paper's contributions and scope?
    \item[] Answer: \answerYes{} % Replace by \answerYes{}, \answerNo{}, or \answerNA{}.
    \item[] Justification: 
    The abstract and introduction clearly state the paper's main contributions and constrain the scope to propagation-dominated photonic devices with localized readouts, evaluated primarily on the 15-wavelength tunable $3\times3$ MMI benchmark. The claims are consistent with the reported main results and ablations, and the paper explicitly notes trade-offs and limited generalization beyond the MMI setting.

\item {\bf Limitations}
    \item[] Question: Does the paper discuss the limitations of the work performed by the authors?
    \item[] Answer: \answerYes{}.
    \item[] Justification: 
    The paper discusses limitations in the Discussion and related experimental sections. It states that the evidence is currently restricted to frequency-domain single-device MMI simulation, that OOD transfer does not show unconditional dominance by PaNO or PaNO-R2, and that NeurOLight remains stronger on some phase-sensitive metrics. It also reports trade-offs introduced by R2 and task-aligned losses, including tail-stability and readout Pareto effects, and provides an efficiency comparison with a caveat that the classical-solver timing is a pipeline measurement rather than a hardware-independent bound.


\item {\bf Theory assumptions and proofs}
    \item[] Question: For each theoretical result, does the paper provide the full set of assumptions and a complete (and correct) proof?
    \item[] Answer: \answerYes{}.
    \item[] Justification: The paper does not introduce formal theorems beyond an elementary discrete port-power bound. The assumptions for this bound are stated in the port-readout definition: a fixed output window, fixed nonnegative port masks, and port power computed as a discrete masked intensity sum on the simulation grid. The bound is numbered and cross-referenced, and it follows directly by substituting the port-power definition and applying the triangle inequality.

    \item {\bf Experimental result reproducibility}
    \item[] Question: Does the paper fully disclose all the information needed to reproduce the main experimental results of the paper to the extent that it affects the main claims and/or conclusions of the paper (regardless of whether the code and data are provided or not)?
    \item[] Answer: \answerYes{}.
    \item[] Justification: The paper provides a reproducibility path for the main experimental claims. The main benchmark, train/test split, input-output representation, model architectures, training budgets, objectives, evaluation metrics, and fixed port-readout protocol are specified in the main text and appendices. code and dataset will be listed in supplementary material together and submit later.

\item {\bf Open access to data and code}
    \item[] Question: Does the paper provide open access to the data and code, with sufficient instructions to faithfully reproduce the main experimental results, as described in supplemental material?
    \item[] Answer: \answerYes{}
    \item[] Justification: code and dataset will be listed in supplementary material together and submit later.

\item {\bf Experimental setting/details}
    \item[] Question: Does the paper specify all the training and test details (e.g., data splits, hyperparameters, how they were chosen, type of optimizer) necessary to understand the results?
    \item[] Answer: \answerYes{} % Replace by \answerYes{}, \answerNo{}, or \answerNA{}.
    \item[] Justification: Section~\ref{sec:experiments} specifies the main 15-wavelength MMI test protocol, grid size, wavelength range, held-out test size, model set, and shared Field/Mediator/Readout evaluator used for all completed rows. Appendix~\ref{app:dataset} gives the dataset construction and deterministic train/test split sizes; Appendix~\ref{app:metrics} defines the reported metrics and readout masks; and Appendix~\ref{app:training} reports the training budgets, objectives, effective batch sizes, and the distinction between 200-epoch main comparisons and 100-epoch ablations.

\item {\bf Experiment statistical significance}
    \item[] Question: Does the paper report error bars suitably and correctly defined or other appropriate information about the statistical significance of the experiments?
    \item[] Answer: \answerYes{}.
    \item[] Justification: The paper reports statistical information for the main experimental comparisons through the supplemental evaluation reports. For the main 15-wavelength MMI benchmark, the shared evaluator reports mean $\pm$ standard deviation over the fixed held-out test cases for the key Field/Mediator/Readout metrics, and paired significance tests for selected main metrics using per-sample errors from the same test split.

\item {\bf Experiments compute resources}
    \item[] Question: For each experiment, does the paper provide sufficient information on the computer resources (type of compute workers, memory, time of execution) needed to reproduce the experiments?
    \item[] Answer: \answerYes{} % Replace by \answerYes{}, \answerNo{}, or \answerNA{}.
    \item[] Justification: Appendix and the supplemental compute audit provide a run-by-run resource summary for the reported experiments. They state the worker type and memory/storage (single NVIDIA RTX 5090 GPU with 32~GB VRAM, Xeon Gold 6459C host CPU, 754~GiB RAM, and dataset/checkpoint storage footprints), wall-clock times for the main 200-epoch comparison, ablations, and LP20+FT30 transfer runs, the forward-pass timing protocol used for the efficiency claim, the total reported training/adaptation budget (about 121.7 GPU-hours), and a separate disclosure of exploratory compute beyond the reported results.

\item {\bf Code of ethics}
    \item[] Question: Does the research conducted in the paper conform, in every respect, with the NeurIPS Code of Ethics \url{https://neurips.cc/public/EthicsGuidelines}?
    \item[] Answer: \answerYes{}.
    \item[] Justification: The paper studies photonic field simulation on synthetic MMI benchmarks and does not involve human subjects, personal data, animal experiments, or other privacy/fairness-sensitive materials. The reported claims are tied to disclosed benchmarks, fixed evaluation metrics, and archived experimental setting. There is no deviation from the NeurIPS Code of Ethics is identified based on the content.

\item {\bf Broader impacts}
    \item[] Question: Does the paper discuss both potential positive societal impacts and negative societal impacts of the work performed?
    \item[] Answer: \answerNA{}.
    \item[] Justification: This work is a technical study of neural surrogate models for synthetic photonic field simulation and is not deployed in a human-facing decision system. It does not use human-subject data, personal data, behavioral data, or sensitive attributes, and it does not produce content or predictions about people. The main risks are technical and engineering-facing, such as inaccurate surrogate predictions in photonic design workflows; these are discussed as limitations and evaluation trade-offs rather than broader societal impacts.
    \item[] Guidelines:
    \begin{itemize}
        \item The answer \answerNA{} means that there is no societal impact of the work performed.
        \item If the authors answer \answerNA{} or \answerNo, they should explain why their work has no societal impact or why the paper does not address societal impact.
        \item Examples of negative societal impacts include potential malicious or unintended uses (e.g., disinformation, generating fake profiles, surveillance), fairness considerations (e.g., deployment of technologies that could make decisions that unfairly impact specific groups), privacy considerations, and security considerations.
        \item The conference expects that many papers will be foundational research and not tied to particular applications, let alone deployments. However, if there is a direct path to any negative applications, the authors should point it out.
        \item The authors should consider possible harms that could arise when the technology is being used as intended and functioning correctly, harms that could arise when the technology is being used as intended but gives incorrect results, and harms following from misuse.
        \item If there are negative societal impacts, the authors could also discuss possible mitigation strategies.
    \end{itemize}

\item {\bf Safeguards}
    \item[] Question: Does the paper describe safeguards that have been put in place for responsible release of data or models that have a high risk for misuse (e.g., pre-trained language models, image generators, or scraped datasets)?
    \item[] Answer: \answerNA{}.
    \item[] Justification: The work does not release or study models or datasets with a high risk of misuse, such as language models, image generators, surveillance systems, or Internet-scraped human data. The models are photonic field surrogates trained on synthetic electromagnetic simulation data, and the primary risks are technical accuracy and engineering validation rather than misuse requiring release safeguards.

\item {\bf Licenses for existing assets}
    \item[] Question: Are the creators or original owners of assets (e.g., code, data, models), used in the paper, properly credited and are the license and terms of use explicitly mentioned and properly respected?
    \item[] Answer: \answerNo{}.
    \item[] Justification: The paper cites the external scientific software and model families used for comparison, but the anonymized submission draft does not yet list explicit software-version and license information for every implementation dependency and generated artifact. The released artifact should include a license file and a dependency/license manifest before camera-ready release.

\item {\bf New assets}
    \item[] Question: Are new assets introduced in the paper well documented and is the documentation provided alongside the assets?
    \item[] Answer: \answerYes{} % Replace by \answerYes{}, \answerNo{}, or \answerNA{}.
    \item[] The anonymized supplementary material includes a self-contained code package with the retained training code, evaluation pipeline, paper and baseline configuration files, and accompanying documentation (\texttt{README.md}, \texttt{DATA\_LAYOUT.md}, and \texttt{LICENSES.md}) describing environment setup, expected data layout, and reproduction commands for the main experiments. The released assets are code and configuration files for simulated photonic experiments; no human-subject consent is required. The supplementary package is anonymized for review.

\item {\bf Crowdsourcing and research with human subjects}
    \item[] Question: For crowdsourcing experiments and research with human subjects, does the paper include the full text of instructions given to participants and screenshots, if applicable, as well as details about compensation (if any)?
    \item[] Answer: \answerNA{}.
    \item[] Justification: The paper does not involve crowdsourcing, user studies, human-subject experiments, participant data, or human annotation labor. The experiments use synthetic photonic simulation data and deterministic evaluation scripts.

\item {\bf Institutional review board (IRB) approvals or equivalent for research with human subjects}
    \item[] Question: Does the paper describe potential risks incurred by study participants, whether such risks were disclosed to the subjects, and whether Institutional Review Board (IRB) approvals (or an equivalent approval/review based on the requirements of your country or institution) were obtained?
    \item[] Answer: \answerNA{}.
    \item[] Justification: The paper does not involve crowdsourcing, human-subject experiments, participant data, or human annotation labor. The experiments use synthetic photonic simulation data, so IRB approval or equivalent human-subject review is not applicable.

\item {\bf Declaration of LLM usage}
    \item[] Question: Does the paper describe the usage of LLMs if it is an important, original, or non-standard component of the core methods in this research? Note that if the LLM is used only for writing, editing, or formatting purposes and does \emph{not} impact the core methodology, scientific rigor, or originality of the research, declaration is not required.
    %this research?
    \item[] Answer: \answerNA{}
    \item[] Justification: LLMs are not part of the core method, model, training pipeline, or evaluation procedure in this work. They were used only as auxiliary tools during project development, such as assisting with code drafting, plotting, and exploratory analysis that helped identify the research problem. The scientific claims, proposed methodology, experiments, and conclusions do not depend on an LLM as an important, original, or non-standard methodological component.

\end{enumerate}

\end{document}